\documentclass[conference]{IEEEtran}
\IEEEoverridecommandlockouts
\usepackage{cite}
\usepackage{amsmath,amssymb,amsfonts}
\usepackage{algorithm}
\usepackage[noend]{algpseudocode}
\usepackage{subfigure}
\usepackage{footnote}
\usepackage{multirow}
\usepackage{booktabs}
\usepackage{enumitem}
\usepackage{amsthm}
\usepackage{graphicx}
\usepackage{textcomp}
\usepackage{xcolor}
\def\BibTeX{{\rm B\kern-.05em{\sc i\kern-.025em b}\kern-.08em
    T\kern-.1667em\lower.7ex\hbox{E}\kern-.125emX}}
    
\def\model{FedConD}

\begin{document}

\title{Asynchronous Federated Learning for Sensor Data with Concept Drift 
}

\author{\IEEEauthorblockN{Yujing Chen}
\IEEEauthorblockA{\textit{Computer Science} \\
\textit{George Mason University}\\
Fairfax, USA \\
ychen37@gmu.edu}
\and
\IEEEauthorblockN{Zheng Chai}
\IEEEauthorblockA{\textit{Computer Science} \\
\textit{George Mason University}\\
Fairfax, USA \\
zchai2@gmu.edu}
\and
\IEEEauthorblockN{Yue Cheng}
\IEEEauthorblockA{\textit{Computer Science} \\
\textit{George Mason University}\\
Fairfax, USA \\
yuecheng@gmu.edu}
\and
\IEEEauthorblockN{Huzefa Rangwala}
\IEEEauthorblockA{\textit{Computer Science} \\
\textit{George Mason University}\\
Fairfax, USA \\
rangwala@gmu.edu}
}

\maketitle

\begin{abstract}
Federated learning (FL) involves multiple distributed devices jointly training a shared model without any of the participants having to reveal their local data to a centralized server. Most of previous FL approaches assume that data on devices are fixed and stationary during the training process. However, this assumption is unrealistic because these devices usually have varying sampling rates and different system configurations. In addition, the underlying distribution of the device data can change dynamically over time, which is known as concept drift. Concept drift makes the learning process complicated because of the inconsistency between existing and upcoming data. Traditional concept drift handling techniques such as chunk based and ensemble learning-based methods are not suitable in the federated learning frameworks due to the heterogeneity of local devices. We propose a novel approach, \model, to detect and deal with the concept drift on local devices and minimize the effect on the performance of models in asynchronous FL. The drift detection strategy is based on an adaptive mechanism which uses the historical performance of the local models. The drift adaptation is realized by adjusting the regularization parameter of objective function on each local device. Additionally, we design a communication strategy on the server side to select local updates in a prudent fashion and speed up model convergence. Experimental evaluations on three evolving data streams and two image datasets show that \model~detects and handles concept drift, and also reduces the overall communication cost compared to other baseline methods. 
\end{abstract}


\section{Introduction}
With the rapid expansion of IoT devices in our digital universe and the explosion in the quantity of data generated by these device sensors, 
on-device learning has emerged as a new paradigm that enables the training of statistical models locally on the devices \cite{bennis2019smartphones,dhar2019device}. Keeping data on device, federated learning (FL) offers an approach to do training locally in a way that a global model is collaboratively trained under the coordination of a central server \cite{konevcny2016federated,mcmahan2017communication,nishio2019client,mills2019communication}.

%
In general, most of the existing FL solutions operate under the assumption that device data are stationary. However, in the real-world, due to the underlying data generating mechanism, the distribution of the data is constantly evolving and the generation of data streams is in the non-stationary environment. This phenomenon is known as concept drift \cite{mehta2017concept,liu2017regional}, which exists commonly in the scenarios of large scale data learning. For example, user behavior/activity prediction models trained before the COVID-19 may not work equally well during COVID-19 pandemic as the user's behaviors changes, weather prediction models change from season to season, customer consumption patterns and underlying recommender systems vary due to the pandemic, economy and so on. 
In other research fields, concept drift has  been referred as covariate shift \cite{sugiyama2012machine}, dataset shift \cite{storkey2009training} and non-stationary learning \cite{gama2014survey}.       

In federated learning process, the occurrence of concept drift leads to  a shift of  test data distribution  from the original training
data leading to serious failures in terms of classification/regression performance\cite{gama2014survey}. Therefore, it is crucial that the federated learning algorithms work adaptively. In general, concept drift can be divided into two types based the changing speed of data patterns: gradual and sudden drift \cite{tsymbal2004problem}.
In case of gradual drift, data distribution changes significantly between the underlying data and the incoming data in a long period, while in sudden drift, this large amounts of change happens in a relatively short period of time. In real-world scenarios, drift could be a mixture of both types \cite{sarnelle2015quantifying}. 
In this paper, we perform study on both these common situations. 

 Previous approaches on coping with concept drift mainly include sliding window approaches, online algorithms, and adaptive ensembles \cite{lu2018learning}.
 Ensemble approaches are popular on improving classification accuracy for concept drift problems \cite{bifet2010leveraging,street2001streaming}.  However, most of these prior work  are deployed in static learning problems and cost more for training. Recent work on asynchronous federated learning  propose an online learning approach to handle the streaming device data \cite{chen2020asynchronous}. While this approach is novel at tackling the challenges associated with heterogeneous edge devices, it cannot deal with the concept drift on local device data. Inspired by this work, we propose a novel asynchronous federated learning method, \model, which can detect and handle concept drift on local devices adaptively. 
 
 In addition, although \cite{chen2020asynchronous} provides dynamic learning strategy on local devices to deal with the varying latency among devices, the global model may still be biased to the devices with lower latency as they communicate more updates to the server. Allowing the server to broadcast the global model to all devices at each round is also a waste of communication bandwidth. In light of these challenges, we design a new strategy which allows the server to send update requests to devices with high latency (e.g., devices with low bandwidth or poor hardware) instead of sending update requests to all devices blindly. This balances the update frequency among devices and leads to a more fair global model. Meanwhile, the overall communication between the server and devices is also reduced with an intelligent selection of local updates.

The main contributions can be summarized as follows.
\begin{itemize}
    \item We propose an asynchronous federated learning framework, FedConD, to detect and handle the data distribution changes across edge devices. 
    \item In this framework the server adopts a novel communication strategy to speedup the model convergence and leads to a more fair global model with balanced frequency of local updates. \model~can improve the predictive performance of the worst $20\%$ devices while also maintains the best test performance for the top  $20\%$ devices. 
    \item Experiments show that the proposed framework achieves better prediction performance than baseline methods with sudden/gradual drift on local devices, and converges faster compared with the other asynchronous online federated learning approach. 
\end{itemize}

\section{Related Work}
\subsection{Federated Learning}
Federated learning  has been proposed as an alternative setting for decentralized learning approaches by McMahan et al. \cite{mcmahan2017communication}: leaving training data distributed on local nodes, a shared global model is learned by aggregating locally-computed updates. Compared to conventional distributed machine learning \cite{ma2017distributed,zhang2015disco,reddi2016aide}, this framework is robust to highly unbalanced and not independent and identically distributed (non-i.i.d) data, unstable network connections and a large number of heterogeneous,  client nodes. Many extensions have been explored based on this original federated learning framework \cite{li2018federated,mills2019communication,chai2020tifl}. 
However, these synchronous FL approaches assume that the device data are stationary, which is not realistic in the real world situation. 
An online learning algorithm with asynchronous communication strategy was proposed by Chen et al.\cite{chen2020asynchronous} to perform federated learning under a more close to real world setting. This solution has no consideration of concept drift during the training process and suffers from the communication bottleneck problem due to the asynchronous framework. To reduce communication costs in federated learning, new settings have been proposed to deal with a federated optimization problem either by model compression or communicating partial model \cite{konevcny2016federated,mills2019communication,jeong2018communication,chai2020fedat}. But the model prediction performance will be hurt by model compression and the convergence speed will be prolonged by partial model communication. 
\subsection{Concept Drift in Non-Distributed Environment}\label{related-work1}
Concept drift is a phenomenon in which the statistical properties of a target domain change over time in an arbitrary manner \cite{lu2014concept}. It was first proposed by Schlimmer et al. \cite{schlimmer1986incremental} who pointed out that noisy data may turn into non-noise information over time. These changes might be caused by changes in hidden variables which cannot be measured directly \cite{liu2017regional}. 

A wide range of algorithms have been developed to address concept drift. According to the changes in data distribution over time, concept drift can be roughly categorized into two types \cite{tsymbal2004problem}: 1) sudden/abrupt drift changing data distribution in a quite short period of time; 2) gradual drift which the large data distribution changes has a relatively longer time. There are some other concept drift types such as incremental drift \cite{lu2016concept}, which has a gradual period of ``intermediate concept''.  The term ``intermediate concept'' was introduced by Gama el al. \cite{gama2014survey} to describe the transformation between concepts. The intermediate concept of gradual drift is one of the starting or ending concept, while in incremental drift the intermediate concept is a mixture of the starting concept and the ending concept \cite{liu2017regional}. In real-world scenarios, drift could be a mixture of many types. In this study, we consider the most common drift types:  (i) sudden drift and (ii) gradual drift. 

Previous work on concept drift uses chunk based learning techniques \cite{kolter2007dynamic,street2001streaming,wang2003mining}, in which a new classifier is learned with a new set of samples. These approaches are not suitable for federated learning, as the sampling rate of device varies, and devices may not generate enough samples quickly enough to build a new classifier. Furthermore, the prediction performance is sensitive to the chunk size, while this parameter is hard to decide in a real-world setting. Other widely used approaches to deal with concept drift problems are ensemble learning-based algorithms, such as using a variant of bagging \cite{bifet2010leveraging}, dynamic adaptation to concept changes (DACC) \cite{jaber2013online}, dynamic weighted majority (DWM) \cite{kolter2007dynamic} and streaming ensemble algorithm (SEA) \cite{street2001streaming}. In these algorithms, a new model is created when drift is detected, but this new model is added to an ensemble pool which also includes older models. These ensemble algorithms are not suitable for on-device learning as they cost more to setup, train, and deploy.

\subsection{Dealing with Concept Drift in Distributed Environment}
Many algorithms have explored machine learning algorithms through parallelization and distribution \cite{zinkevich2009slow,mcdonald2010distributed,zinkevich2010parallelized,agarwal2012distributed,iutzeler2013asynchronous,aybat2015asynchronous}. 
There are very few approaches that address issues related to concept drift in a fully distributed network. Ang et al. \cite{ang2010classifying} propose a P2P learning framework for concept drift classification, which includes a drift detection (reactive behavior) and simultaneously a drift prediction (proactive behavior) mechanisms. The basic idea behind the approach is the use of chunk-based technique with a triggering and an ensemble based approaches. As stated in Section \ref{related-work1}, ensemble approaches are not suitable for on-device learning, and moreover, this approach also suffers from high communication cost for the network wide model propagation. Heged{\H{u}}s el al. \cite{hegedHus2013massively} handle concept drift by maintaining new as well as old models in the network, and models of the data perform random walks in the P2P network. 
While these techniques can be used in distributed learning, the network topology without a central server is fundamentally different from the federated learning framework in this study. 

To the best of our knowledge, currently there are no robust solutions to tackle concept drift efficiently in asynchronous federated learning. We put forward the concept drift issue in the federated learning framework and propose techniques to detect and adapt the concept drift on local devices. Our proposed algorithm is detailed in Section \ref{con-drift}.

\section{Problem Description}
In this section, we define the classical federated learning objective and introduce the definition of concept drift.

\subsection{Preliminaries: Classical Federated Learning}
For classical federated learning algorithms, the goal is to minimize the following objective function:

\begin{equation}
    \min_w \left\{F(w) = \sum_{k=1}^{K}p_k f_k(w).\right\}
\end{equation}

where $K$ is the total number of devices, $p_k \geq 0$ is the weight of $k$-th device and $\sum_{k=1}^{K}p_k = 1$. Suppose $\mathcal{D}_k$ is data captured on device $k$, and let $n_k = |\mathcal{D}_k|$ be the number of samples on device $k$. We can set $p_k$ to be $\frac{n_k}{N}$ , where $N = \sum_k n_k$ is the total number of samples over the entire dataset. The local
objective of client $k$ is defined as:
\begin{equation}
f_k(w_k) \overset{def}= \frac{1}{n_k}\sum_{i\in\mathcal{D}_k}^{} \ell_i(x_i,y_i;w_k).
\end{equation}

where $\ell_i(x_i,y_i;w_k)$ is the corresponding loss function\footnote{Cross-entropy loss for the classification problems and mean absolute error for the regression problems.} for data sample $\{x_i,y_i\}$ and $w_k$ is the local model parameter.

\subsection{Concept Drift}

The distribution $\mathcal{D}_k$ may change over time. For a given time period $[0,t]$, a set of samples $S_{0,t} = ({d_0,...,d_t})$ are from $\mathcal{D}_k$, where $d_t = (x_t,y_t)$ is the data sample at time step $t$, $x_t$ is the feature vector, $y_t$ is the target label. Assume each training example on device $k$ is generated by a source $\mathcal{S}_k$, then a sudden drift occurs when $\mathcal{S}_k^t$ is suddenly replaced by $\mathcal{S}_k^{t+1}$ at time step $t$. Gradual drifts change with a slower rate and roughly can be divided into two types. The first type has a mixed distribution of $\mathcal{D}_k$ and $\mathcal{D}_k'$ during the transition phase. As time goes on, the probability of observing examples from $\mathcal{D}_k$ decreases, while that of examples from $\mathcal{D}_k'$ increases. The second type also be referred as incremental drift \cite{liu2017regional}, which has more than two data distributions, however the difference between any adjacent distributions is small and the transition usually takes a longer period of time. 


Concept drift is critical to online learning problems, because such inequality may lead to a inconsistency in decision boundaries, thereby increasing the error rate. Research into concept drift adaptation focuses on how to minimize the drop in accuracy and achieve the fastest recovery rate during the concept transformation process. 
We consider of both of these concept drift types and aim to minimize the performance drops of the learned models in this paper.  

\begin{figure}[t]
\centering
\includegraphics[trim={1cm 0cm 9cm 1.5cm}, clip,width=8.5cm]{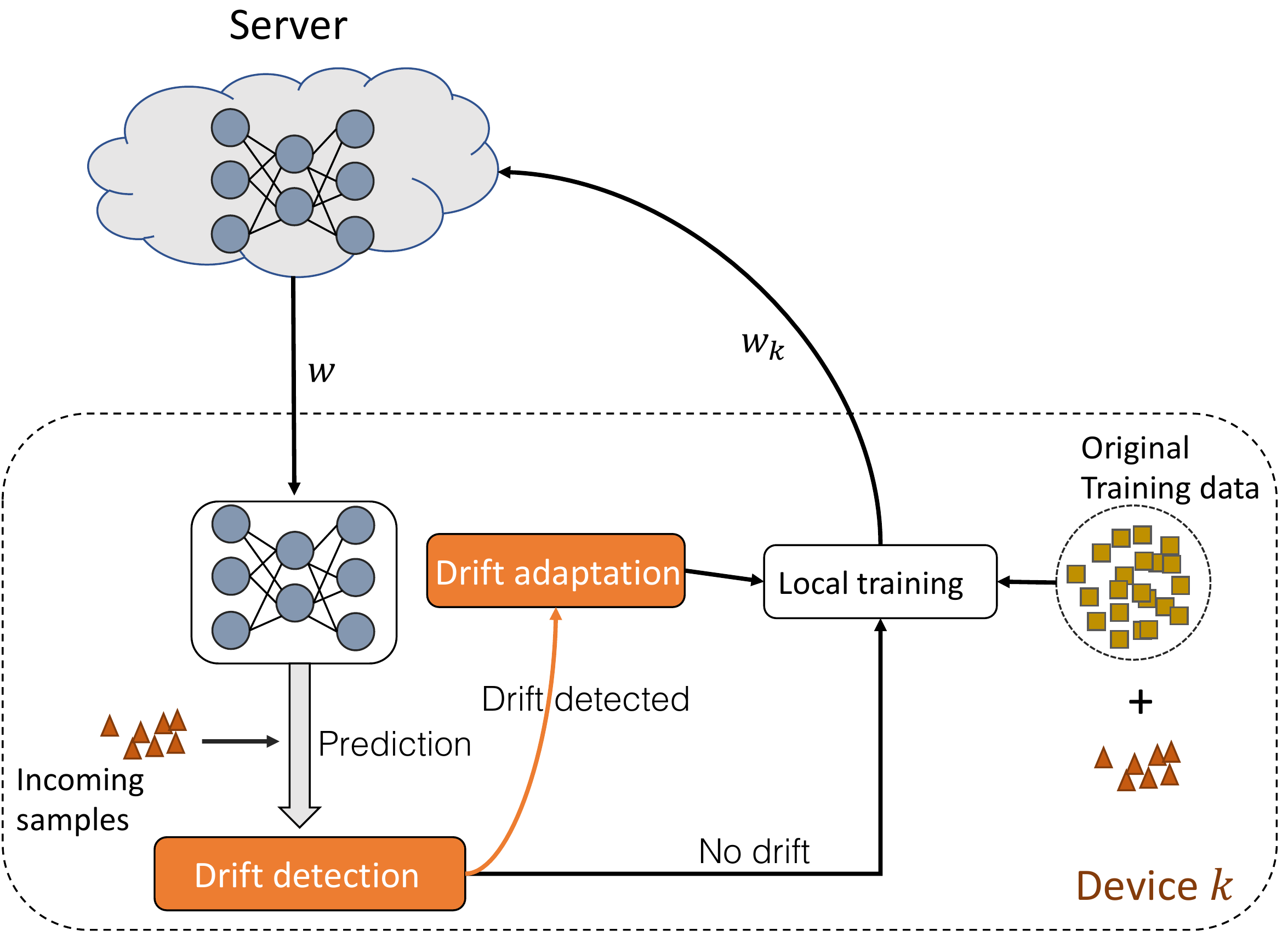}
	\caption{Illustration of the learning process on device $k$. After receiving the global model from server, local device first performs prediction on the new coming samples, then detecting whether there is concept drift. If drift happens, a drift adaptation strategy will be applied before the local training. $w$ is used to represent the global model, and $w_k$ is the trained model of device $k$. 
	}\label{fig:FLcon}
\end{figure}

\section{Method} \label{con-drift}
In this section, we first introduce our strategy for concept drift detection and method to handle the concept drift on local devices. Then we explain the learning procedure on server. 
To begin with, we first update the local objective function of client $k$:

\begin{equation}\label{local-obj}
    h_k(w_k) = f_k(w_k) + \frac{\lambda}{2}||w_k - w||^2
\end{equation}
where $w$ is the global model on the server. The regularization component $\frac{\lambda}{2}||w_k - w||^2$ aims to restrict the amount of local deviation by penalizing large changes from the current model at the server \cite{li2018federated,chen2020asynchronous}. By restricting
the local updates to be closer to the initial (global) model, we can limit the impact of variable
local updates efficiently. 

Figure \ref{fig:FLcon} illustrates the learning process on local device. The device starts to do prediction on the new incoming data after receiving the global model $w$ from server. Next a drift detection process will be applied on the prediction results to determine whether a concept drift happens (detail explanation is in Section \ref{drift_detect}). If a drift is detected, the drift adaptation process is initialized before the training starts, otherwise the training will start directly. Finally the trained local model $w_k$ will be uploaded to server for aggregation.  

\begin{algorithm}[t]
\caption{Algorithm of \model}\label{alg:FLconcept}
\begin{algorithmic}[1]
    \State \textbf{Input:} $K$ distributed devices with related data distribution,
  regularization parameter $\lambda$, threshold $\delta$, parameter $\gamma$.
 \State \textbf{Initialize:} a portion of $\gamma$ devices to be available for training.
\State \underline{\textbf{Learning at Server}} 
  \For{rounds $t=1,2,...,T$} 
       \State  /* get the update on $w^t$ */
        \State
        compute $w^t$ \Comment{[Eq.(\ref{equa:server})]} 
       \State  broadcast $w^t$ to the device with the fewest updates  
   \EndFor
   \State  \textbf{end for}
\State \underline{\textbf{Learning Process of Local Client $k$ at round $t$}}
     \State receive $w^t$ from the server
    
     \State $s_k^{a + 1}$ $\leftarrow$ evaluate on new incoming samples
     \State $\Bar{s_k} = mean[s_k^1, s_k^2,...,s_k^{a}]$ 
     \State $\hat{s_k} = mean[s_k^1, s_k^2,...,s_k^{a},s_k^{a+1}]$
     \State Perform statistical testing and get $\Gamma_k$, then use $\Gamma_k$ to get the corresponding p-value $P$
     \Comment{[Eq.(\ref{drift-detection})]} 
     \If{$P > significance \: level$}
     \State continue
     \Else
     \State increase $\lambda$
     \EndIf

 \State Update $w_k^{t+1} \leftarrow w_k^t-\eta_k^t \nabla h_k $
 \State upload $w_k^{t+1}$ to the server
\end{algorithmic}
\end{algorithm}

\subsection{Local Drift Detection}\label{drift_detect}
The main idea behind our concept drift detection algorithm is that we can use the new generated samples on the devices for evaluating the models before we use them for training. Then using the results got from this evaluation we can decide whether the concept drifted. 

For device $k$, at time step $t$ the classifier predicts the class label of $x_{t+1}$ to be $\hat{y}_{t+1}$. Most of previous works assume that the true label $y_{t+1}$ is provided after some time, and both the $y_{t+1}$ and $\hat{y}_{t+1}$ are available for the drift detection\cite{brzezinski2013reacting,ang2010classifying,hegedHus2013massively,lu2014concept}, which is, the learning is in a supervised framework. Then, example $x_{t+1}$ with its label $y_{t+1}$ becomes a part of the training data and the process is repeated when the next instance is observed. 

 We perform drift detection using statistical testing. Each time we do prediction on the new incoming data with the downloaded server model, and then perform evaluation on the predicted labels and the true labels. Next we add the evaluation result into a queue with bounded size. Suppose device $k$ has performed local training $a$ times, and let $s_k$ be the evaluation value, then the evaluation queue is $[s_k^1, s_k^2,...,s_k^{a}]$. By storing these evaluations in the history, we can detect the trend of the performance of the model and decide whether there is a drift occurs at time $a + 1$. We assume that $s_k^{a+1}$ equals to $\Bar{s}_k = \mu(s_k^1, s_k^2,...,s_k^{a})$ if no concept drift happens; and a significant decrease of $s_k^{a+1}$ suggests that the
concept is changing.  The test is performed by calculating the following statistic\footnote{For computational efficiency, we use the Yates's correction for the Pearson chi-square test.}:
\begin{equation}\label{drift-detection}
   \Gamma_k =  \frac{|\Bar{s}_k - s_k^{\alpha+1}|-0.5\Delta_k}{\sqrt{\hat{s}(1-\hat{s})\Delta_k}}
\end{equation}

where $\Delta_k = \frac{1}{a} + 1$ and $\hat{s} = \mu(s_k^1, s_k^2,...,s_k^{a},s_k^{a+1})$. We compare its value to the percentile of the standard normal distribution
to obtain the observed significance level (p-value). If the p-value, $P$, is less than a significance level, then the null
hypothesis ($s_k^{a+1} = \Bar{s}_k$) is rejected and the alternative hypothesis ($s_k^{a+1} > \Bar{s}_k$) is accepted, namely concept drift has been detected. We set the significance level as 0.05 (5\%), which can indicate a difference is significant if $P$ is less than this value. The bounded size of the queue is set to be $20$ for our cases.

\subsection{Local Drift Adaptation}\label{drift_handle}
The added regularization component in Equation (\ref{local-obj}) is a penalty term to the loss function.
For instance,  Zhang el al. \cite{zhang2014deep} propose elastic averaging SGD in deep networks with a similar penalty term in its objective. In distributed methods, Shamir et al. propose DANE \cite{shamir2014communication} and Reddi et al. provide AIDE \cite{reddi2016aide} use similar regularization term in the local objective function, while these methods are within the data center setting. Li el al. propose FedProx \cite{li2018federated} to add this penalty term to tackle heterogeneity in federated networks, and they show that FedProx is robust to systems heterogeneity with stable convergence compared to vanilla FedAvg. 

In Equation (\ref{local-obj}), $\lambda$ controls the amount of penalty added to the original local objective $f_k(\cdot)$. When a drift is detected, the local model will deviate further from the global model, thus we should increase the weight of penalty component to force the local model to be close to the global model. We show empirically that this strategy can safely incorporate variable amounts of
local work resulting from concept drift. 

\subsection{Learning on Server}
The server aggregates the received local models in an asynchronous manner, that is, server model will be updated after receiving one device's update. Suppose at round $t+1$, server receives the update from device $k$. Let $w^{t+1}$ be the server model, $w_k^t$ be the local model, $n_k^{t+1}$ be the number of samples on devices $k$ and $N^{t+1}$ be the total samples over all devices. The server model is updated as follow: 
\begin{equation}\label{equa:server}
\begin{aligned}
    w^{t+1} &= w^t - \frac{n_k^{t+1}}{N^{t+1}}(w_k^t - w_k^{t+1})\\
    &= w^t -\frac{n_k^{t+1}}{N^{t+1}}( w_k^t - (w_k^t - \eta_k^t \nabla h_k(w^t))) \\
    &= w^t - \eta_k^t\frac{n_k^{t+1}}{N^{t+1}} \nabla h_k(w^t). 
\end{aligned}
\end{equation}
where $\eta_k$ is the learning rate of device $k$, and $\nabla h_k(\cdot)$ is the gradient on the local data of device $k$. 
\subsection{Communication-efficient Learning}
Previous asynchronous FLs have no constrains on local device selection, that is, the server broadcasts global model $w$ to all available devices at each round \cite{xie2019asynchronous,chen2020asynchronous}. This communication strategy is not efficient due to 1) downlink communication cost is high due to the server need to communicate with all local devices at each round; 2) the server can easily become a communication bottleneck with tens of thousands of devices updating the model simultaneously; 3) the update frequency of devices vary due to reasons like system heterogeneity, network bandwidth or data heterogeneity. Therefore, the aggregated global model may bias to devices which have more frequent updates. 

In former work of communication-efficient federated learning, one commonly used approach to reduce the overall communication cost is compressing the model size transferred between the server and devices \cite{mills2019communication,jeong2018communication,reisizadeh2019exact}. While these compression techniques can reduce the communication cost efficiently, the model performance is also hurt because the whole model information cannot be preserved during the compression-decompression process. Different from the synchronous FL frameworks, which allow only a portion of devices to perform local training and communicate with the server. With large amount of local devices, asynchronous FL framework should select local device update more wisely and also control the number of local devices which perform training simultaneously. To address the above mentioned problems we propose a communication-efficient strategy to reduce the overall communication cost during the learning process in asynchronous FL frameworks, and furthermore, learn a more fair global model with faster convergence speed. At the server side we maintain records of the update frequency of all devices $\{p_1,p_2,...,p_K\}$. At the beginning of each round, the server will send update request to the device with fewest updates, which is, getting $min (p_1,p_2,...,p_K)$ and send server model to the according device. Meanwhile, we design a parameter $\gamma$ to control the number of local devices which are performing local training at the same time. A detailed analysis on this parameter can be found in Section \ref{tune-parameter}.

\begin{table*}[]
\small
\centering
\caption{Prediction performance and statistics of all methods on four datasets. 'Avg' is the average prediction performance on all devices; 'Var' shows the variance of the final prediction performance distribution of devices. We show the results of Air Quality data on $20\%$ devices with concept drift due to this dataset has a smaller number of devices. }\label{acc_statsall}
\begin{tabular}{l|llllll}
\hline
&  & \begin{tabular}[c]{@{}l@{}}Cifar-10\\ (Accuracy$\uparrow$)\end{tabular} & \begin{tabular}[c]{@{}l@{}}Fashion-Mnist\\ (Accuracy$\uparrow$)\end{tabular} & \begin{tabular}[c]{@{}l@{}}FitRec\\ (Smape $\downarrow$)\end{tabular} & \begin{tabular}[c]{@{}l@{}}Air Quality\\ (Smape $\downarrow$)\end{tabular} & \begin{tabular}[c]{@{}l@{}}ExtraSensory\\ (F1-score$\uparrow$)\end{tabular} \\ \hline
\multirow{4}{*}{Avg (all devices)}  & \multicolumn{1}{l|}{FedAvg}  & 0.822       & 0.913  & 0.897 & 0.446 & 0.527\\
& \multicolumn{1}{l|}{FedProx} & 0.841     & 0.912 & 0.857  & 0.443 & 0.567\\
& \multicolumn{1}{l|}{ASO-Fed} & 0.892      & 0.943 & 0.832  & 0.470 & 0.711\\
& \multicolumn{1}{l|}{\model(proposed)}   & \textbf{0.907}     & \textbf{0.953}  & \textbf{0.822}  & \textbf{0.430} & \textbf{0.721} \\ 
\hline
\hline
\multirow{4}{*}{Var (all devices)} & \multicolumn{1}{l|}{FedAvg}  & 1.65e-05    & 2.31e-05 & 0.0768 & 1.75e-3 & 0.0475\\
& \multicolumn{1}{l|}{FedProx} & 2.42e-04   & 6.79e-04  & 0.0894 & 2.80e-3 & 0.0369\\
& \multicolumn{1}{l|}{ASO-Fed} & 1.58e-05  & 4.21e-06 & 0.0668 & 6.75e-04 & 0.0261\\
& \multicolumn{1}{l|}{\model(proposed)}   & \textbf{1.09e-05}  & \textbf{2.38e-06} & \textbf{0.0671} & \textbf{2.51e-04} & \textbf{0.0226}\\ 
\hline
\multirow{4}{*}{Var (drift devices)} & \multicolumn{1}{l|}{FedAvg}  & 3.59e-7 & 1.10e-4 & 1.57e-1 & 9.05e-4 & $-$\\
& \multicolumn{1}{l|}{FedProx} & 0.2e-05    & 3.10e-3 & 5.40e-4 &  2.92e-3 & $-$ \\
& \multicolumn{1}{l|}{ASO-Fed} & 0.39e-4  & 0.4e-05  & 8.40e-4 & 2.01e-4 &$-$ \\
& \multicolumn{1}{l|}{\model(proposed)}   & \textbf{0.85e-07} & \textbf{0.4e-05}  & \textbf{1.80e-4}  &  \textbf{0.11e-4} &$-$ \\
\hline
\end{tabular}
\end{table*}

\begin{figure*}[htbp]
  \centering
  \subfigure[Cifar-10 (Accuracy $\uparrow$)]{\includegraphics[scale=0.27]{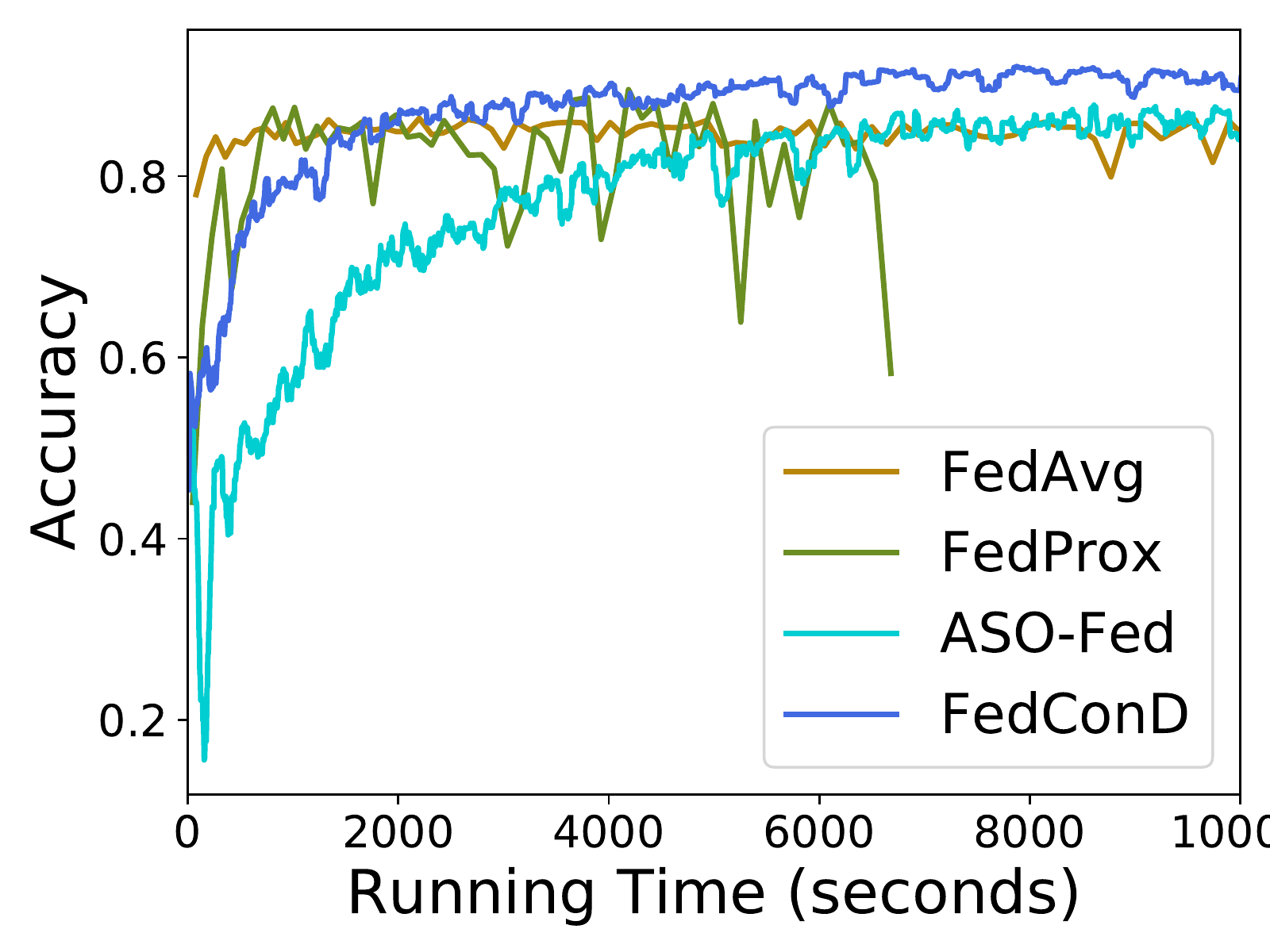}}
  \subfigure[Fashion-MNIST (Accuracy $\uparrow$)]{\includegraphics[scale=0.27]{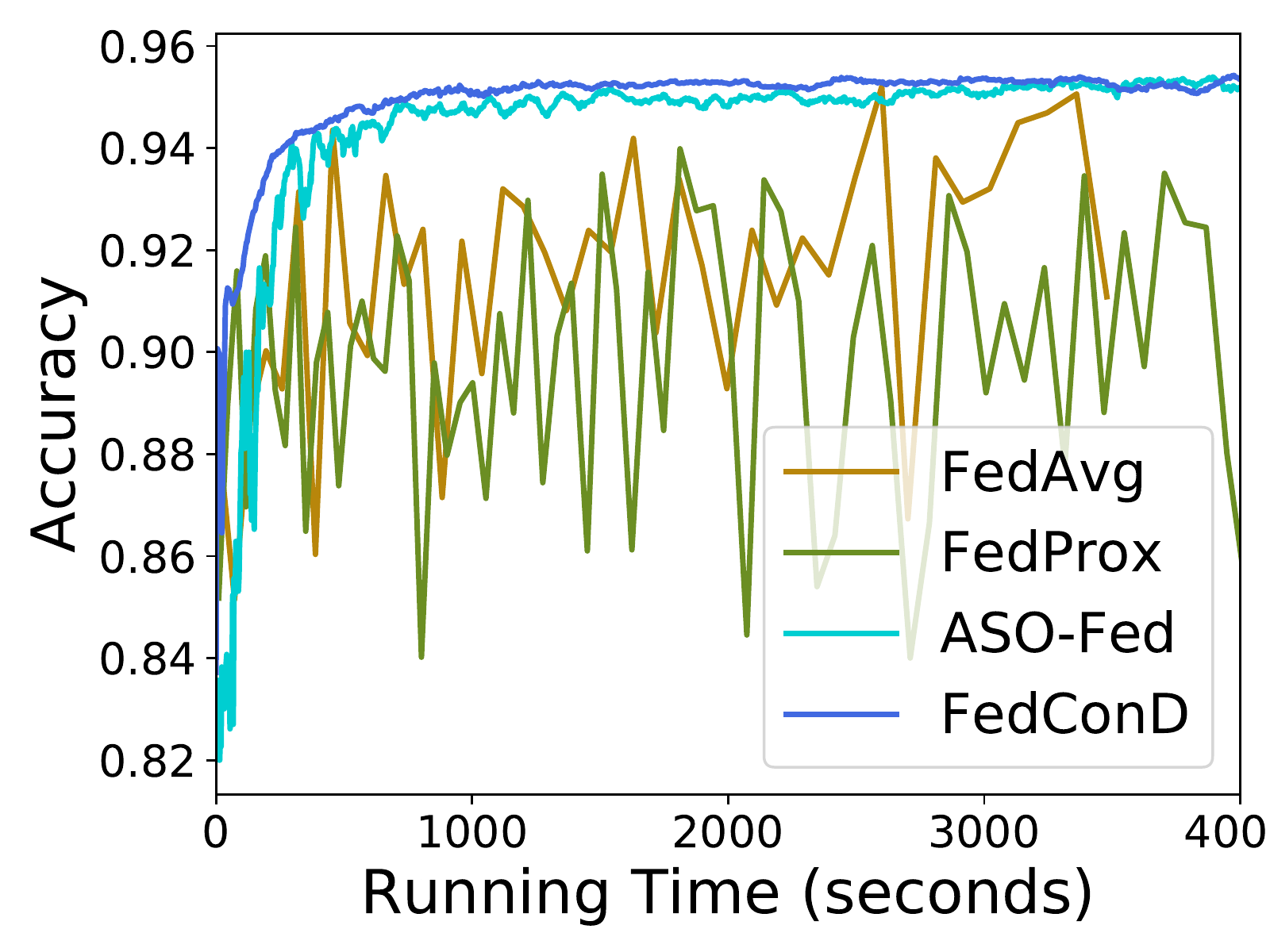}}
  \subfigure[FitRec (SMAPE $\downarrow$)]{\includegraphics[scale=0.27]{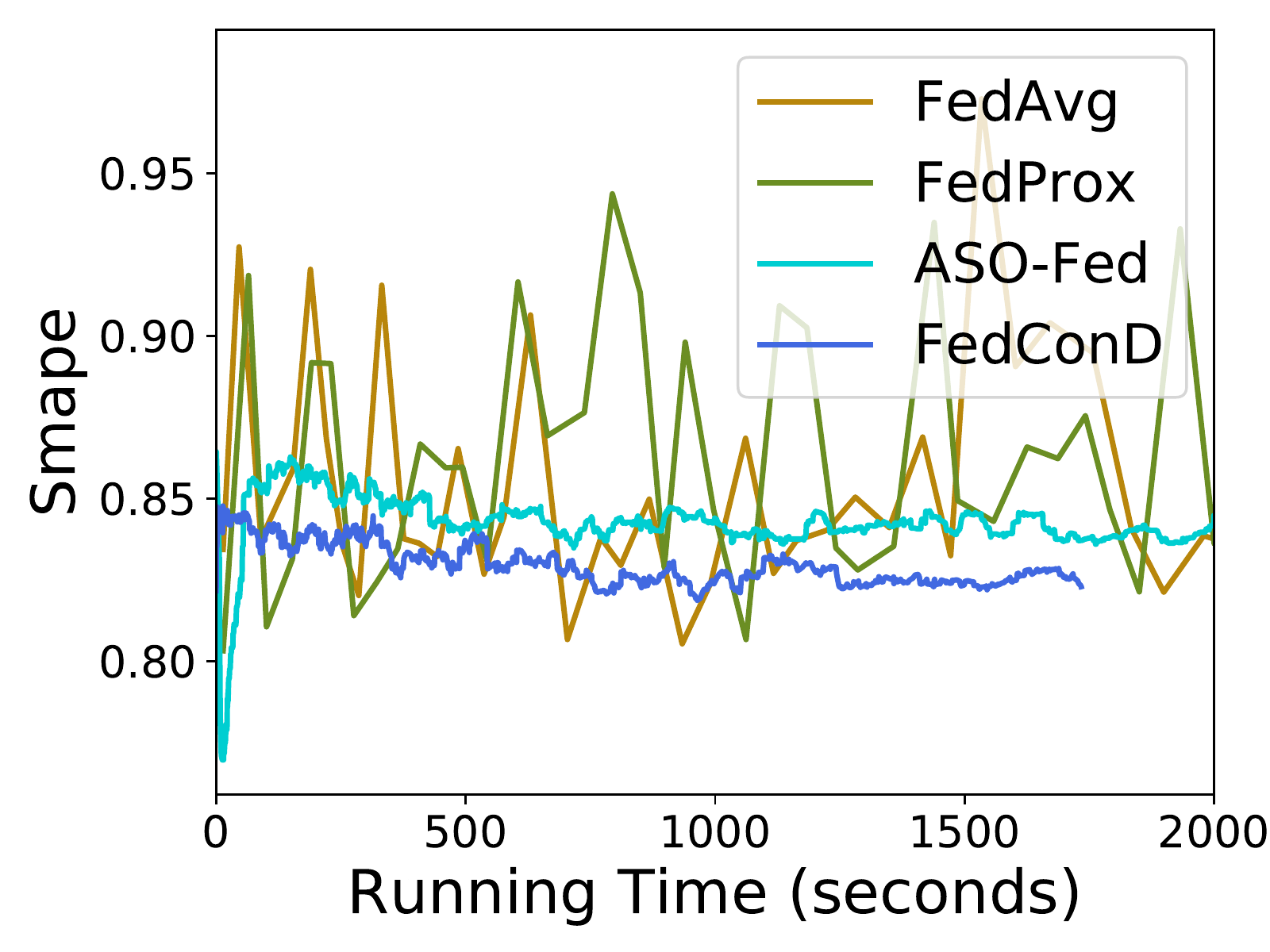}}
  \subfigure[Air Quality (SMAPE $\downarrow$)]{\includegraphics[scale=0.27]{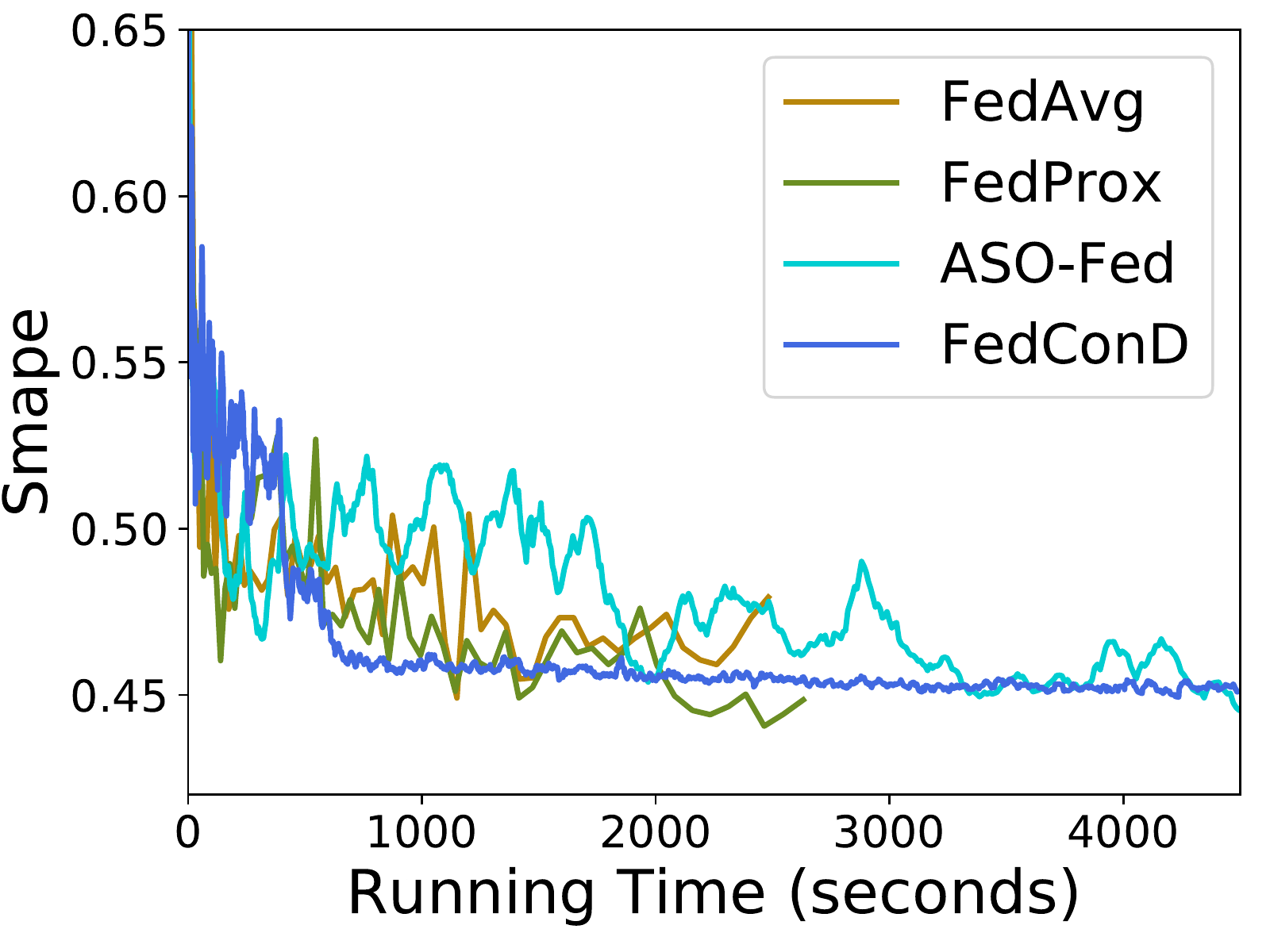}}
  
  \caption{Test set performance vs. running time for four datasets. Lower SMAPE value indicates better model performance. For the synchronized federated frameworks, we plot results of \textit{FedAvg} and \textit{FedProx} at every $5$ global iterations. }\label{tab:prediction-curve}
\end{figure*}

\section{Experimental Evaluation}
We perform extensive evaluations on three evolving data streams and two image datasets. 
\subsection{Datasets}

\begin{itemize}

\item \textbf{FitRec Dataset\footnote{https://sites.google.com/eng.ucsd.edu/fitrec-project/home}:} 
    User sport records generated on mobile devices and uploaded to Endomondo, including multiple sources of sequential sensor data such as heart rate, speed, GPS as well as the sport type (e.g., biking, hiking), user gender and weather condition. Following \cite{ni2019modeling}, we re-sampled the data in 10-second intervals. We use data of randomly selected 30 users for heart rate prediction. 
	\item \textbf{Air Quality Dataset\footnote{https://biendata.com/competition/kdd\_2018/data/}:} Air quality data collected from multiple weather sensor devices distributed in 9 locations of Beijing from Jan 2017 to Jan 2018, with features such as thermometer and barometer. Each area is modeled as a separate participant and the observed weather data is used to predict the measure of six air pollutants (e.g., PM2.5, PM10) from May 1st, 2018 to May 31st, 2018. 
	\item \textbf{ExtraSensory Dataset \footnote{http://extrasensory.ucsd.edu/}:} Mobile phone sensor data (e.g., location services, audio, accelerator) collected from 60 users, performing any of 51 activities (e.g., walking, talking, running), using personal smart phones/watches was used as a more realistic alternative to the generated data set to show how the proposed algorithm could be applied in a different context. \cite{vaizman2018extrasensory}.
	We use the provided 225-length feature vectors of time and frequency domain variables generated for each instance. 
	
    \item \textbf{Fashion-MNIST\footnote{https://research.zalando.com/welcome/mission/research-projects/fashion-mnist/}:} This is a dataset of Zalando's article images—consisting of a training set of 60,000 examples and a test set of 10,000 examples. Each example is a 28x28 grayscale image, associated with a label from 10 classes (e.g., Dresses, Coat, Bag). Each class has the same number of examples. We follow a non-IID setting as in \textit{FedAvg} \cite{mcmahan2017communication} and divide the data into 20 parts according to their labels. We first sort the data by category label, divide each category into 4 different sizes $\{2000,2750,3250,4000\}$, and assign each of 20 parts 2 different sizes. We model each part as a separate client and predict the target labels.
    
    \item \textbf{Cifar-10\footnote{https://www.cs.toronto.edu/~kriz/cifar.html}:} This dataset contains 60,000 images associated with a label from 10 classes (e.g., airplane, dog, cat). Each class has the same number of examples. There are 50,000 training samples and 10,000 test samples. Each image is a 32x32 colour image. We follow a non-IID setting as in \textit{FedAvg} \cite{mcmahan2017communication} and divide the data into 20 parts according to their labels. We first sort the data by category label, divide each category into 4 different sizes $\{1500,2250,2750,3500\}$, and assign each of 20 parts 2 different sizes. We model each part as a separate client and predict the target labels.

\end{itemize}

\begin{table*}[t]
\small
\centering
\caption{Statistics of the test performance for proposed \model on Cifar-10 and FitRec data. \model~can improve the test performance (Avg) on the bottom $20\%$ devices without hurting the test performance on the top $20\%$ devices. The variance (Var) of the final performance distribution maintains the lowest value of \model.}\label{statistics}
\begin{tabular}{l|llll|llll}
\hline
                       & \multicolumn{4}{c|}{Cifar-10(Accuracy$\uparrow$)}                                              & \multicolumn{4}{c}{FitRec(Smape$\downarrow$)}                                                \\ \hline
\multirow{2}{*}{Model} & \multicolumn{2}{c|}{Top 20\%}           & \multicolumn{2}{c|}{Bottom 20\%} & \multicolumn{2}{c|}{Top 20\%}           & \multicolumn{2}{c}{Bottom 20\%} \\ \cline{2-9} 
                       & Avg & \multicolumn{1}{l|}{Var} & Avg        & Var        & Avg & \multicolumn{1}{l|}{Var} & Avg        & Var       \\ \hline
FedAvg                 & 0.823   & 2.62e-04                      & 0.817          & 6.11e-04        & 0.381   & 1.42e-02                      & 1.07           & 0.92e-03       \\
FedProx                & 0.830   & 0.13e-04                      & 0.821          & 0.12e-04        & 0.668   & 6.93e-02                      & 0.932          & 1.35e-02       \\
ASO-Fed                & 0.899   & 0.61e-05                      & 0.868          & 3.51e-06        & 0.624   & \textbf{1.10e-02}                      & 0.879          & 6.57e-02       \\
\model                  & \textbf{0.906}   & \textbf{0.34e-05}                      & \textbf{0.889}          & \textbf{7.32e-07}        & \textbf{0.601}   & 1.21e-02                      & \textbf{0.831}          & \textbf{0.55e-03}       \\ \hline
\end{tabular}
\end{table*}

\subsection{Comparative Methods} 
We compare the proposed approach with the following synchronous and asynchronous federated learning approaches.

\begin{itemize}
    \item FedAvg \cite{mcmahan2017communication,konevcny2016federated}: the commonly used synchronous federated learning approach proposed by McMahan {\em et al.} \cite{mcmahan2017communication}.
    \item FedProx \cite{li2018federated}: synchronous federated learning framework with a proximal term on the local objective function to mitigate the data heterogeneity problem and to improve the model stability compared to FedAvg.
    \item ASO-Fed \cite{chen2020asynchronous}: asynchronous federated learning framework to deal with non-IID streaming device data.
\end{itemize}

\begin{figure*}[htbp]
  \centering
  \subfigure[Cifar-10 (Accuracy $\uparrow$)]{\includegraphics[scale=0.27]{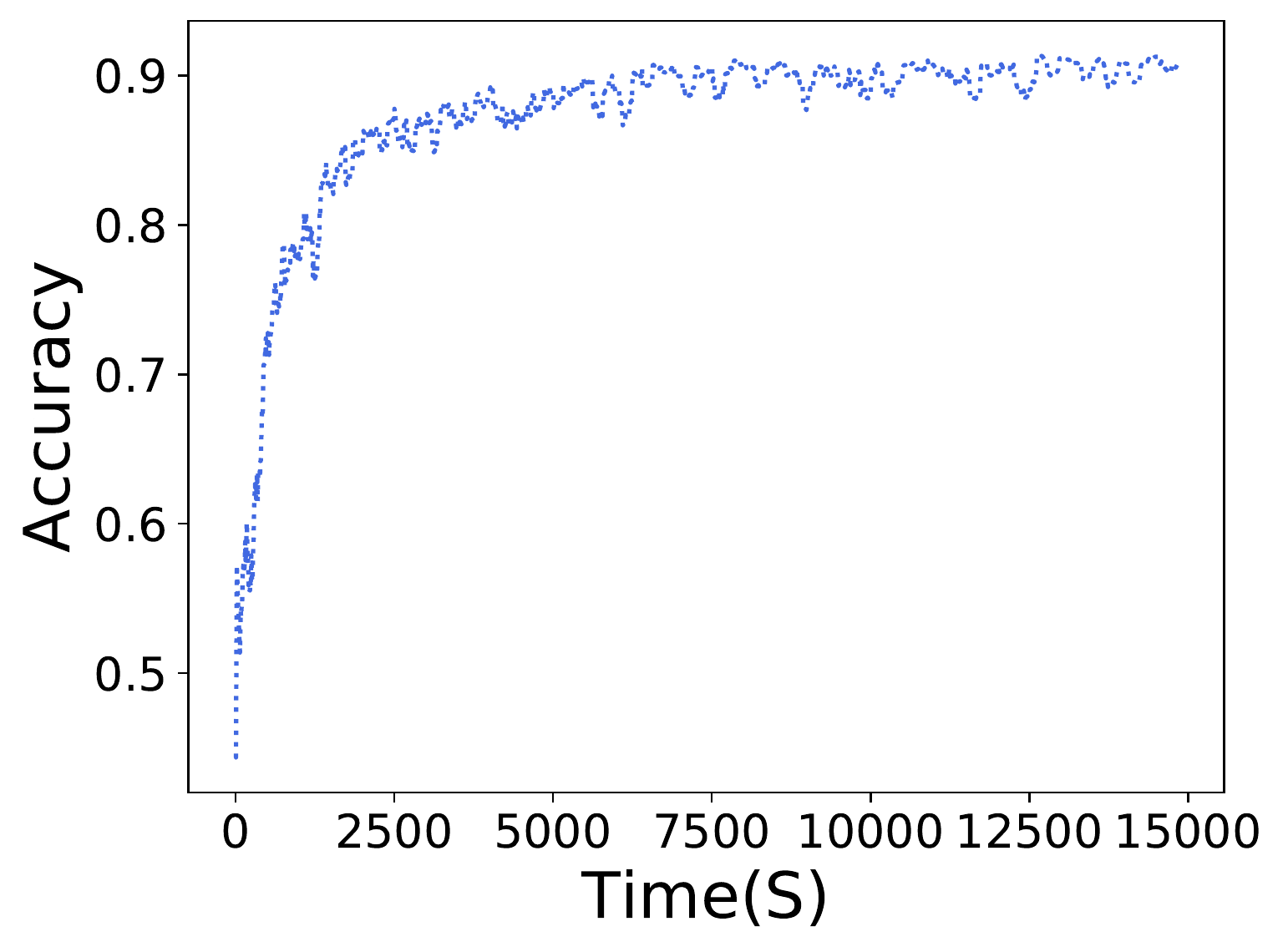}}
  \subfigure[Fashion-MNIST (Accuracy $\uparrow$)]{\includegraphics[scale=0.27]{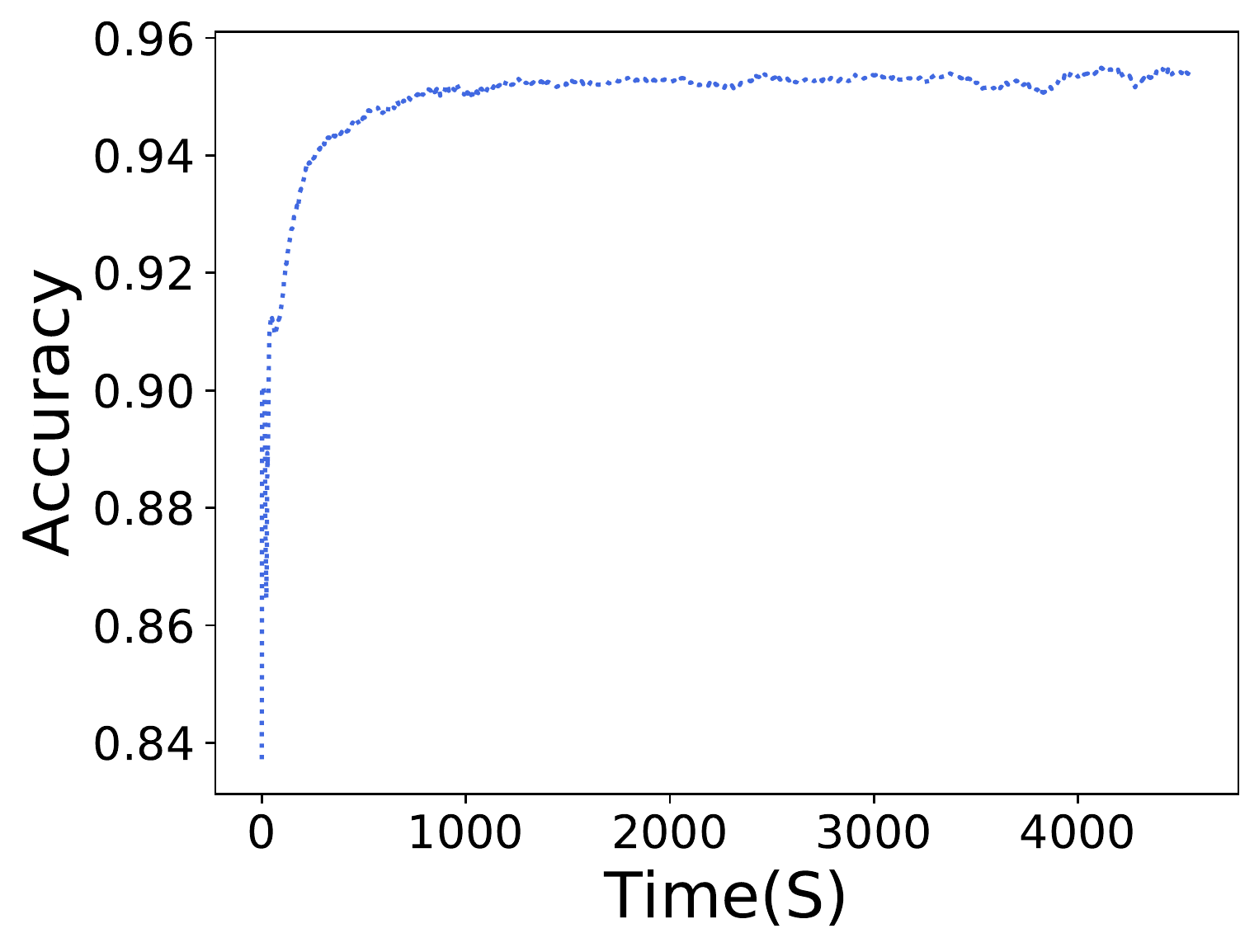}}
  \subfigure[FitRec (SMAPE $\downarrow$)]{\includegraphics[scale=0.27]{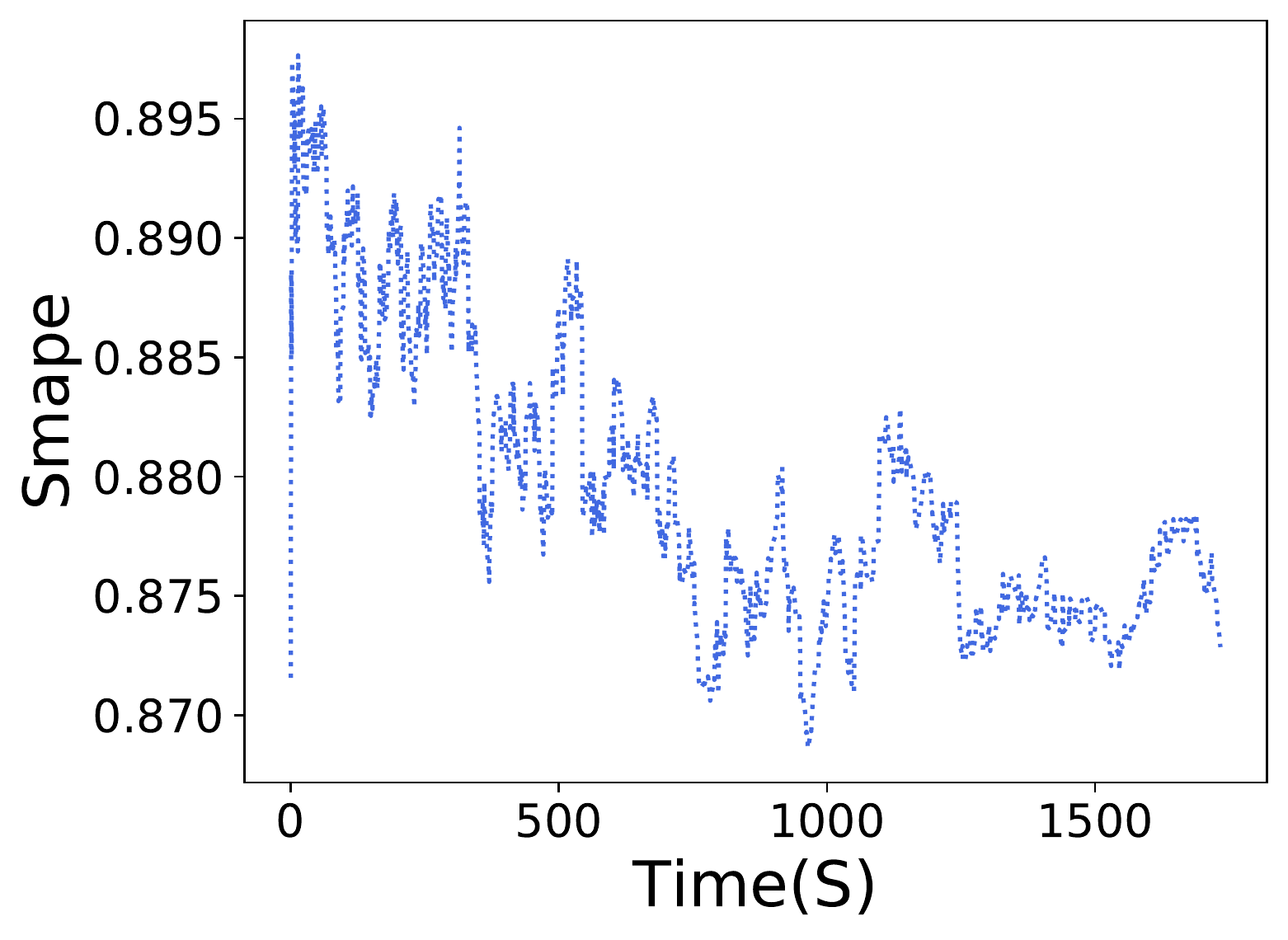}}
  \subfigure[Air Quality (SMAPE $\downarrow$)]{\includegraphics[scale=0.27]{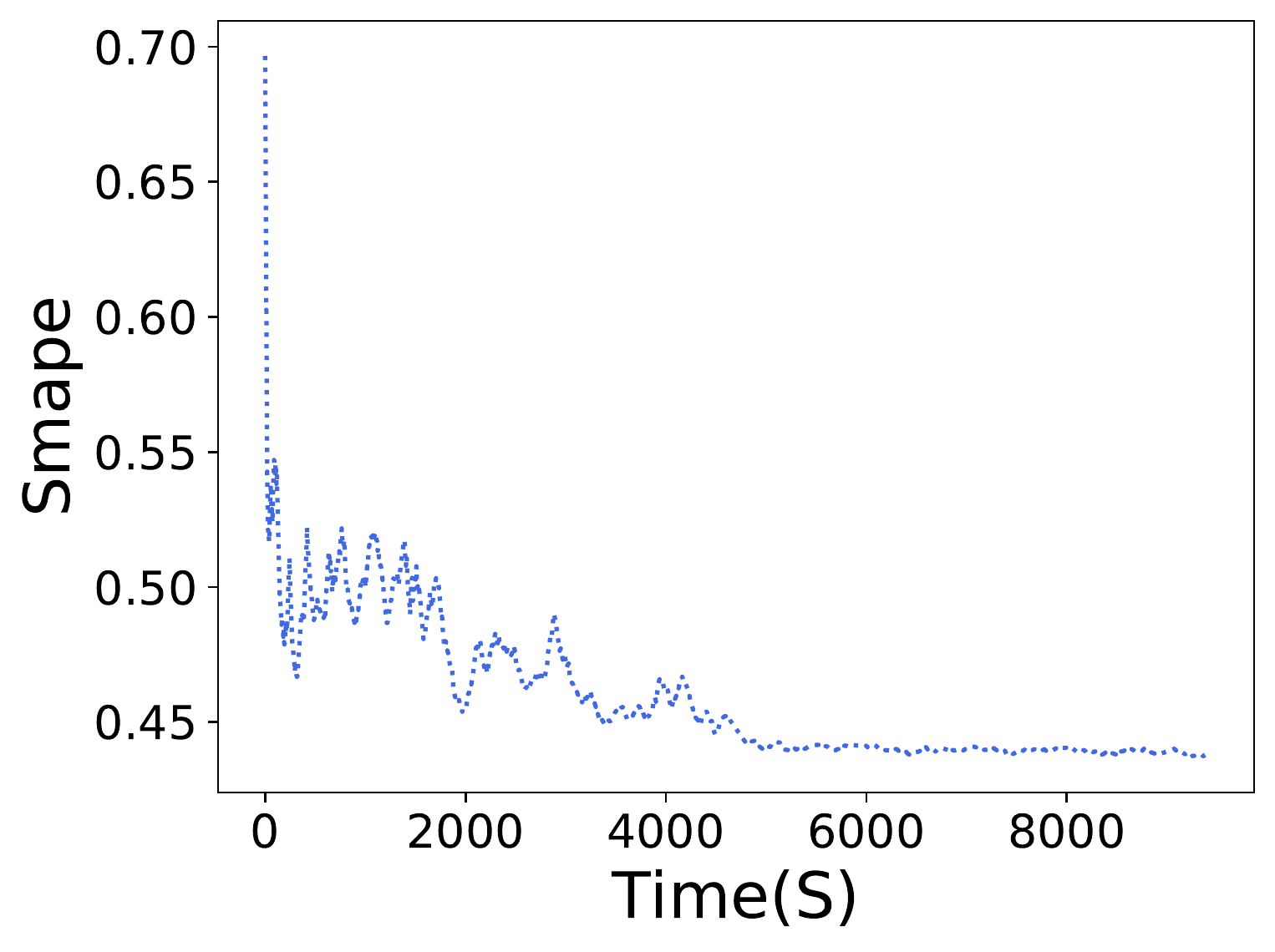}}
  
  \caption{Server model evaluation performance of \model~on four datasets. We plot results of \textit{FedAvg} and \textit{FedProx} at every $5$ global iterations. }\label{tab:model-curve}
\end{figure*}

\subsection{Experimental Setup}
\subsubsection{Drift Simulation}\label{drift-simulation}
 
Real-world data satisfy the gradual drift situation as these data are generated by sensors in the non-stationary environment. For instance, data on each weather sensor of the Air Quality Dataset changes from season to season. Thus we use the three provided real-world datasets to test the algorithm for gradual drift. To make the problem more challenging, following \cite{vzliobaite2009determining}, we simulate the sudden concept drift in time-series datasets. First we randomly select a portion $C$ of devices to add concept drifts. Then we modify the feature values to be random values between $10-1000$ of a consecutive time period of samples and keep the according labels unchanged. We perform this sudden concept drift simulation on both the FitRec dataset and Air Quality Dataset.

There are drawbacks for real-world data for evaluating concept drift handling methods, as the precise start and end time of drifts is unknown. 
Due to these limitations, it is difficult to evaluate methods and understand the drift \cite{lu2018learning}. Thus we also simulate concept drifts on the image data. We use the similar strategy as Mansukhani et al. \cite{mansukhani2019}, which is, adding noise to a portion of the image data. Same as the time-series datasets, we randomly select a portion $C$ of participants to add noise. We set $C$ as $0.1$, which is, there are $10\%$ devices with concept drifts in each dataset. To further evaluate the model performance with this parameter changes, we report the results with different values of $C$ in Section \ref{tune-parameter}.

\subsubsection{Training Details}
We split the each device's data into $60\%$, $20\%$, $20\%$ for training, validation, and testing, respectively. As for each training data, we start with a random portion of the total training size, and increase by $0.01\%-0.1\%$ each iteration to simulate the arriving data. We set 
the threshold $\delta$ value is $0.2$ for image classification data and $0.25$ for time-series regression data. We set the portion of participation devices of FedAvg and FedProx as $0.2$ at each round. We use a single layer LSTM followed by a fully connected layer for the two time-series datasets and two CNN layers followed by a max pooling layer for the two image datasets. The local epoch number of each client is set as $2$. We design simple network architectures for all datesets so that it can be easily handled by mobile devices. All of the experiments are conducted with two Intel E5-2660 v3 10-core CPU at 2.60GHz \cite{duplyakin2019design}.

\begin{figure}[htbp]
  \centering
  \subfigure[Fashion-MNIST (Accuracy $\uparrow$)]{\includegraphics[scale=0.27]{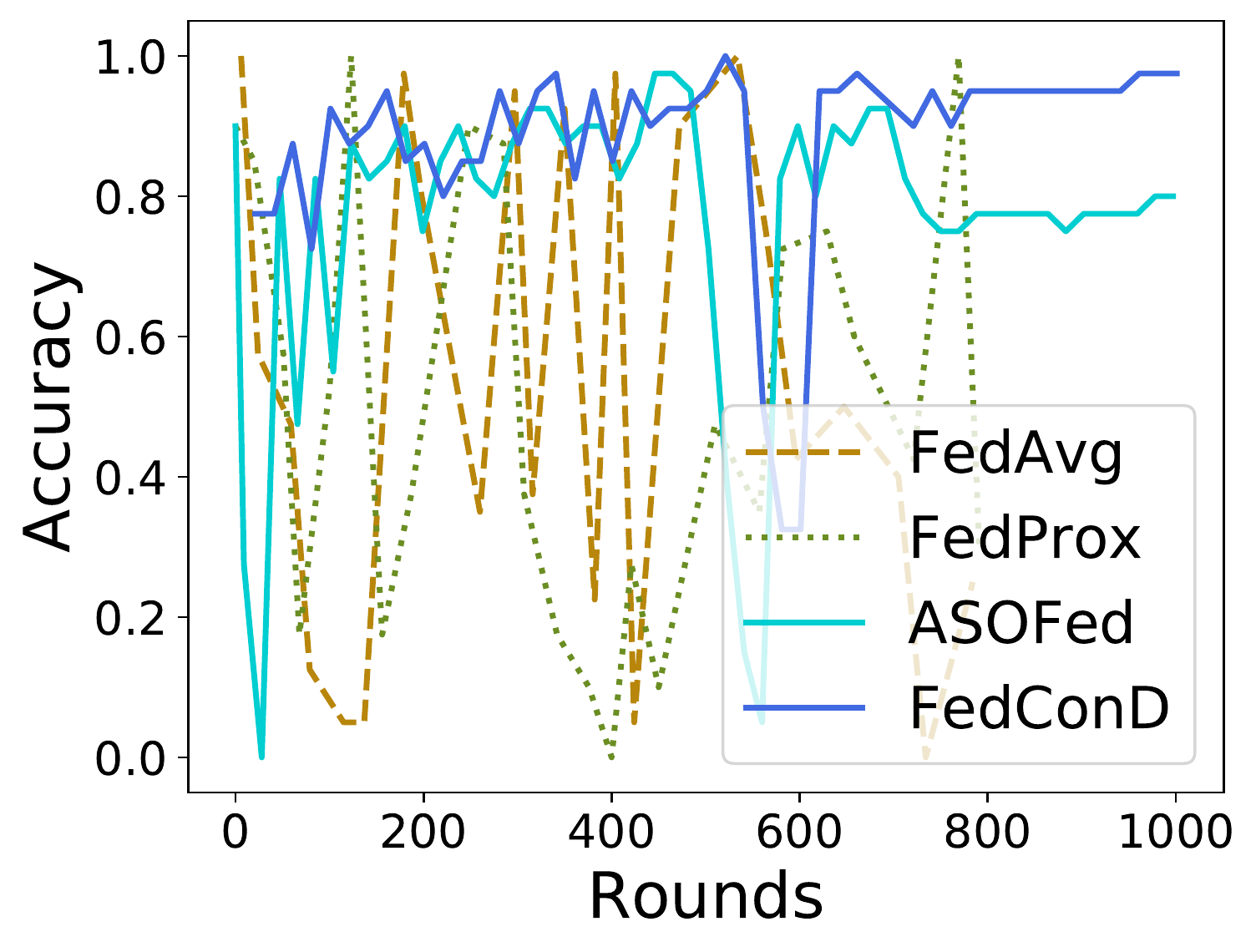}}
  \subfigure[Fashion-MNIST (Accuracy $\uparrow$)]{\includegraphics[scale=0.27]{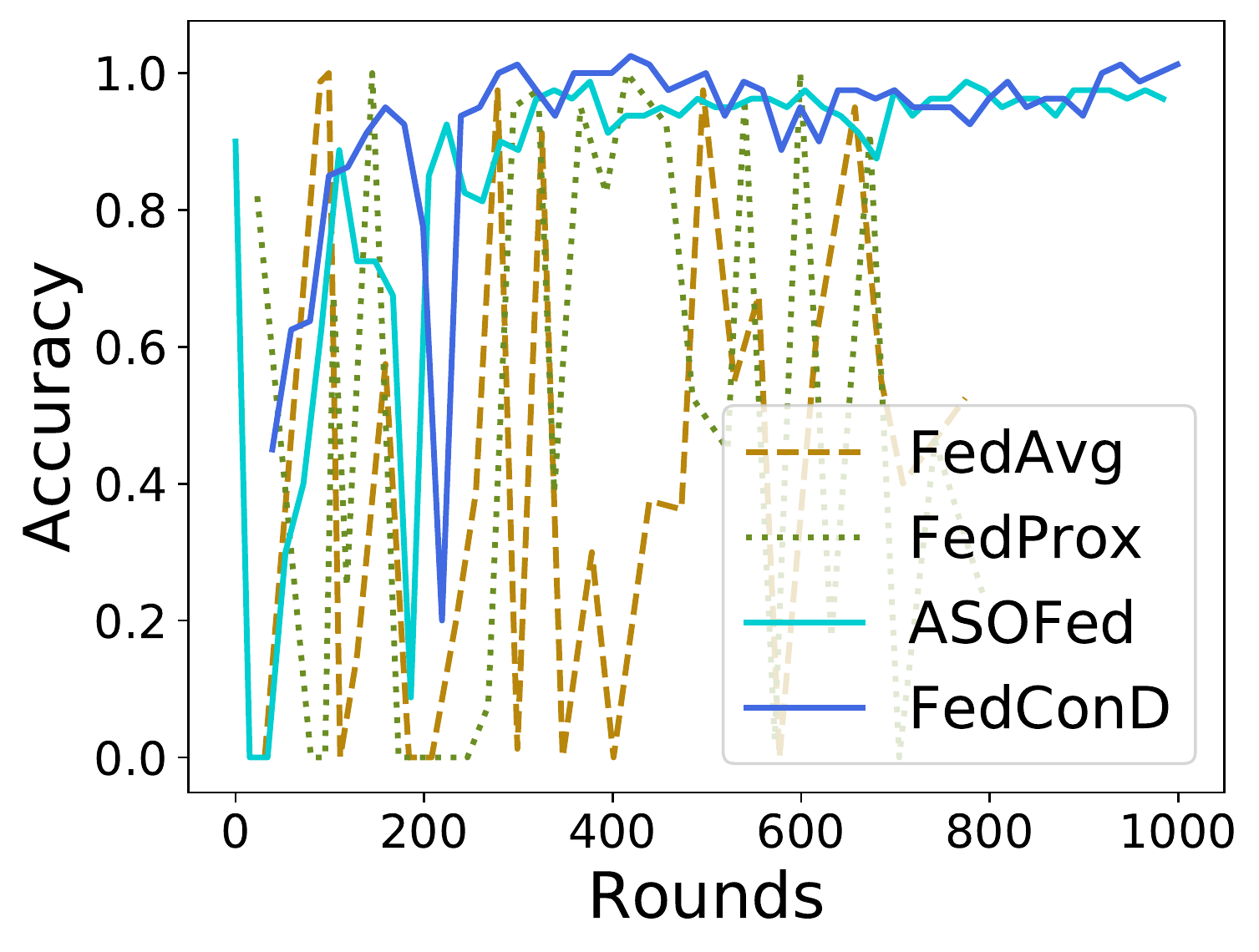}}
  
  \caption{Evaluation performance of local devices which have concept drift of Fashion-MNIST.}\label{tab:drift-curve}
\end{figure}

\begin{table}[]
\centering
\caption{\textit{Total Bytes} required for upload(Uplink) and download (Downlink) to achieve a certain target performance on different learning tasks. Note the model size within the same learning tasks is the same across all methods. }\label{communication-cost}

\begin{tabular}{l|l|ll}
\hline                                                                                       & Model    & \multicolumn{1}{l}{ASO-Fed} & \model \\ \hline
\multirow{2}{*}{\begin{tabular}[c]{@{}l@{}}Cifar-10\\ (Acc.: 0.75)\end{tabular}}         & Uplink   & 15124.9 MB                          & 3708.2 MB                   \\
                                                                                         & Downlink & 302321.8 MB                         & 14832.5 MB                   \\ \hline
\multirow{2}{*}{\begin{tabular}[c]{@{}l@{}}Fashion-MNIST\\ (Acc.: 0.94)\end{tabular}}    & Uplink   & 3173.7 MB                          & 2386.1 MB                    \\
                                                                                         & Downlink & 63462.4 MB                         & 9547.5 MB                   \\ \hline
\multirow{2}{*}{\begin{tabular}[c]{@{}l@{}}Air Quality\\ (SMAPE: 0.46)\end{tabular}}     & Uplink   & 20.14 MB                          & 13.88 MB                    \\
                                                                                         & Downlink & 199 MB                          & 27.76 MB                 \\ \hline
\multirow{2}{*}{\begin{tabular}[c]{@{}l@{}}ExtraSensory\\ (F1-score: 0.60)\end{tabular}} & Uplink   & 698.6 MB                        & 574.4 MB                   \\
                                                                                         & Downlink & 41910.3 MB                        & 6893.2 MB                  \\ \hline
\end{tabular}

\end{table}

\section{Experimental Results}
\subsection{Performance Comparison}
Table \ref{acc_statsall} reports the prediction performance comparing the proposed model with the baseline approaches. We observer that \model~receives the best predictive performance (Avg) and the lowest variance (Var) among all the FL approaches. These results indicate that \model~performs well with concept drift on local devices. As the server dynamically requests updates from local devices based on their update frequency, our framework can balance the updates among all devices and learn a more fair optimal global model. We notice that ASO-Fed has the close performance as the proposed model both on the predictive performance and the statistic results. ASO-Fed designs an online learning algorithm with a decay coefficient to balance the previous and current local gradients, which also works on dealing with concept drifts. However, with a fixed regularization parameter across all devices, ASO-Fed is not flexible enough to handle drifts efficiently. Synchronous FLs (FedAvg and FedProx) do not perform well compared with the other two asynchronous approaches. 

To evaluate the proposed drift adaption strategy, we also report the statistics on devices with drifts in Table \ref{acc_statsall}. Each device in ExtraSensory data contains mixed drift types, thus we only report the variance across all devices. 
The results indicates that the proposed model can get the lowest variance within the drift devices.

\begin{figure*}[htbp]
  \centering
  \subfigure[Early Stage Drift]{\includegraphics[scale=0.30]{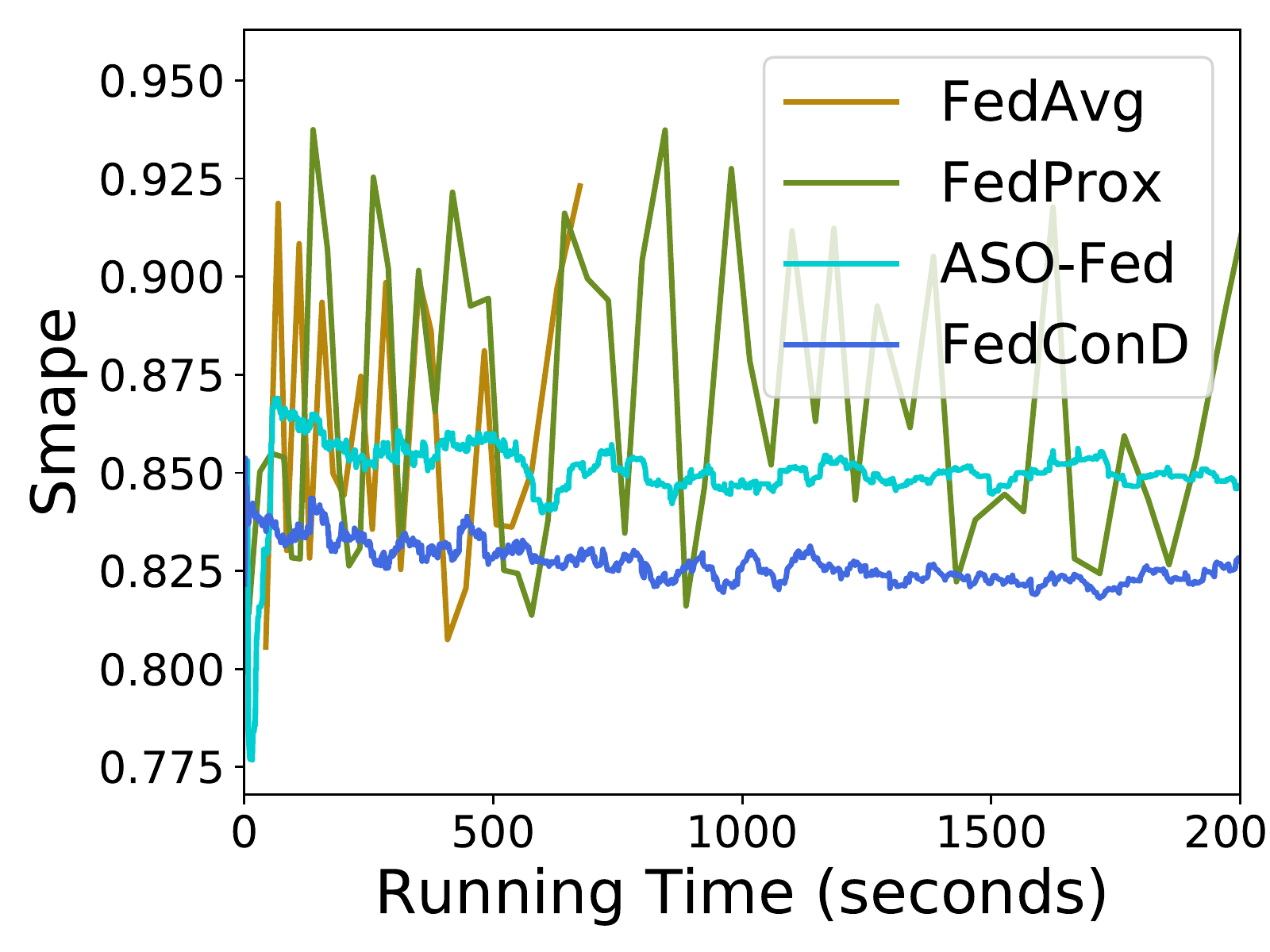}}
  \subfigure[Middle Stage Drift]{\includegraphics[scale=0.30]{charts/smape_FitRec.pdf}}
  \subfigure[Late Stage Drift]{\includegraphics[scale=0.30]{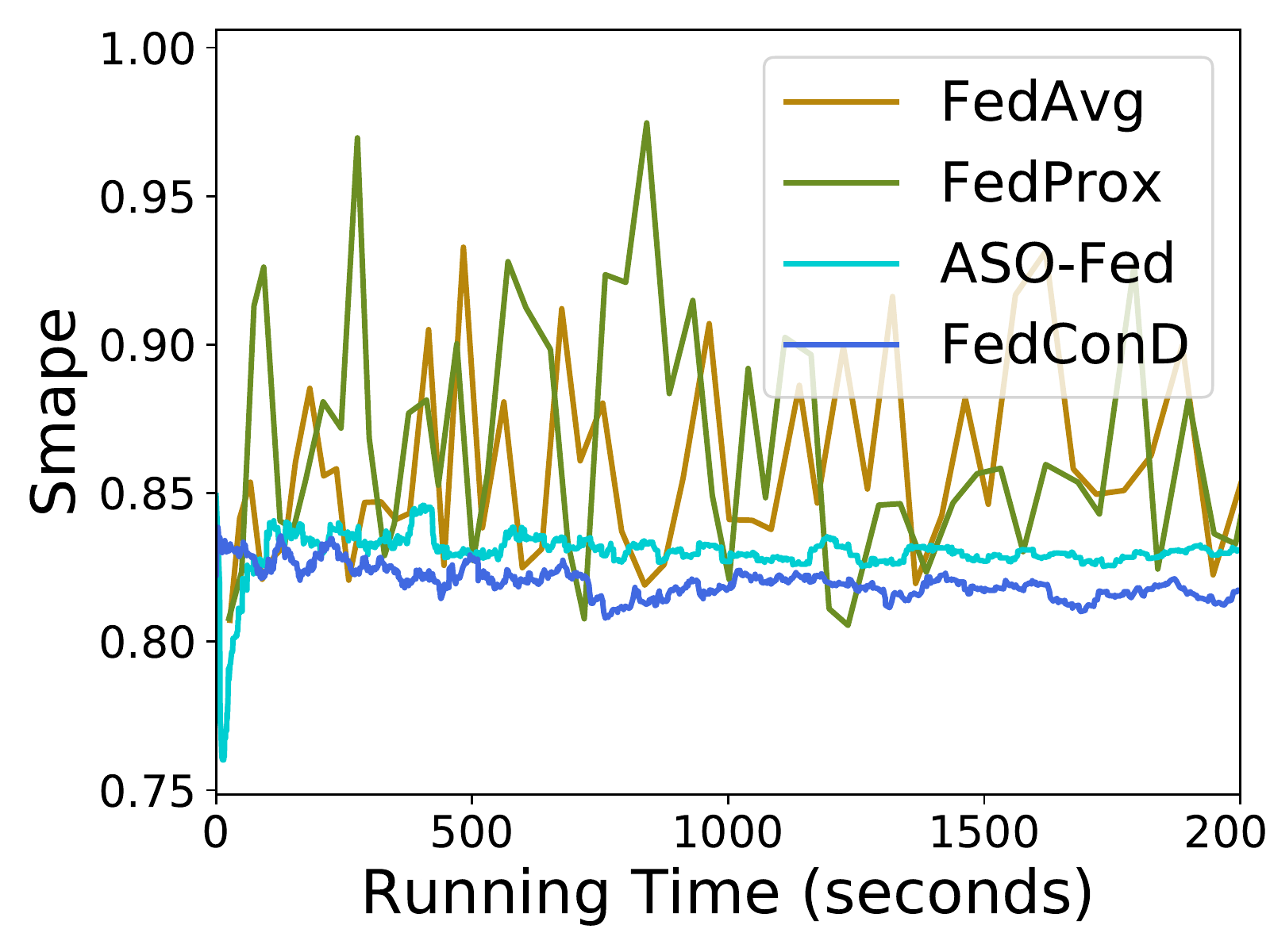}}
  \caption{Convergence of \model~compared with baseline methods for concept drift at 'Early Stage (10\%)', 'Middle Stage (40\%)' and 'Late Stage (70\%)' on the FitRec dataset. }\label{tab:drift-stage}
\end{figure*}

\subsubsection{Communication Efficiency}
Figure \ref{tab:prediction-curve} shows the learning curves of the global model on all FL frameworks. For Cifar-10 and Fashion-MNIST image data classification tasks, our proposed model achieves the best predictive performance and also converges faster than the other three approaches. For FitRec and Air Quality time-series data regression tasks, we observe larger fluctuations on ASO-Fed at the initial learning phase. Due to the high non-IID nature of these two datasets, it is common to see this fluctuations in asynchronous update frameworks as the server aggregates local updates one at each time. 
Our proposed framework \model~ converges more quickly and sees stable performance. The two synchronous frameworks, however, suffer from unstable model performance and cannot handle streaming data.



We further compare the two asynchronous FL methods with respect to the \textit{communicated bytes} required to achieve a certain target accuracy on a federated learning task. Table \ref{communication-cost} 
%
shows the amount of upstream and downstream communications required to achieve the target prediction performance for the asynchronous FL methods. \model manages to achieve the desired target performance within the smallest upload communication budget compared with ASOFed. 
Meanwhile, we see much smaller communication costs required for \model~ for the downstream communications. This is due to the design of $\gamma$ to 
control the number of 
local training devices. We further discuss the selection of $\gamma$ in Section \ref{tune-parameter}.


\subsubsection{Fairness of Client Participation}

The server model aims to get a  equal contributions from the  local clients by adjusting the update frequency of local models. While the FL developer may have different fairness criteria, achieving a more fair accuracy distribution across devices is mostly the paramount \cite{kairouz2019advances}. We follow the popular metrics to measure the fairness of model accuracy for different strategies \cite{kairouz2019advances,li2019fair}.
We evaluate the statistical performance of the test accuracy for \model~on the best $20\%$ and worst $20\%$ devices. As shown in Table \ref{statistics}, \model~can improve the predictive performance of the worst $20\%$ devices while also maintains the best test performance for  the top $20\%$ devices. This indicates that the proposed communication-efficient strategy on the server can lead to a balanced and fair model.
Lacking methods to deal with streaming data and concept drift, FedAvg and FedProx are observed to have large gaps in  the average performance between the top devices and  bottom devices on the studied benchmarks.

\subsection{Analysis on Concept Drift}

\subsubsection{Adaptation to new concepts.} We simulate sudden drift on the image benchmark data by injecting noise on local devices (experimental protocol described in Section \ref{drift-simulation}).
As shown in Figure \ref{tab:prediction-curve}, \model~achieves the best classification performance on Cifar-10 and Fashion-MNIST data compared with other competitors 
It is also noticeable that FedAvg and FedProx fail to converge in this scenario. 
ASO-Fed has the similar accuracy trend as \model, but it needs longer time to converge. This is because ASO-Fed has extra computation such as feature learning component on the server. Streaming data usually contains mixed type of concept drifts. 

\subsubsection{Adaptation to mixed concepts.} We also add sudden drift to the FitRec and Air Quality data to create the complicate drift types with a mix of gradual drift and sudden drift.
%
We notice that the prediction performance of all FL algorithms have larger fluctuations with the mixed drifts. ASO-Fed has noticeable unstable performance at the beginning stage of the training process. FedAvg and FedProx still show no convergence trend. \model~converges to the best performance with the lowest prediction errors.   

To better evaluate the effect of concept drift on both the global model and local models, we further show the predictive performance of the global model for \model~in Figure \ref{tab:model-curve} and local models for \model~in Figure \ref{tab:drift-curve}. As shown in Figure \ref{tab:model-curve}, FitRec is a  non-IID dataset.  Each local device contains data of only one sport type (e.g., hiking, biking). Thus the unstable learning curve is not only due to concept drifts but also caused by the non-IID aspects of this data. 
We observe the obvious drop of the prediction performance on Fashion-MNIST at the initial learning stage. This dataset is simpler than Cifar-10 for it contains only grayscale images, thus the changes on local data distributions can be reflected on the global model. In contrast, drifts on local data cannot be easily noticed on the server side for the Cifar-10 data. Air Quality data has a smaller number of devices, changing even one device's data distribution will have relatively longer affect on the global model.

We show the predictive performance at each round on the drifted devices for the Fashion-MNIST dataset  in Figure \ref{tab:drift-curve}. Obvious drops can be noticed from the learning curve of \model, which indicate the data distribution changes on these local devices and the proposed model can detect these changes. ASO-Fed has similar learning trend as \model~as it has a  local decay strategy to deal with online learning, while \model~has higher accuracy when drift occurs and converges to better performance. FedAvg and FedProx  fail to detect the local drift. 

\begin{figure}[htbp]
  \centering
  \subfigure[Fashion-MNIST (Accuracy $\uparrow$)]{\includegraphics[scale=0.26]{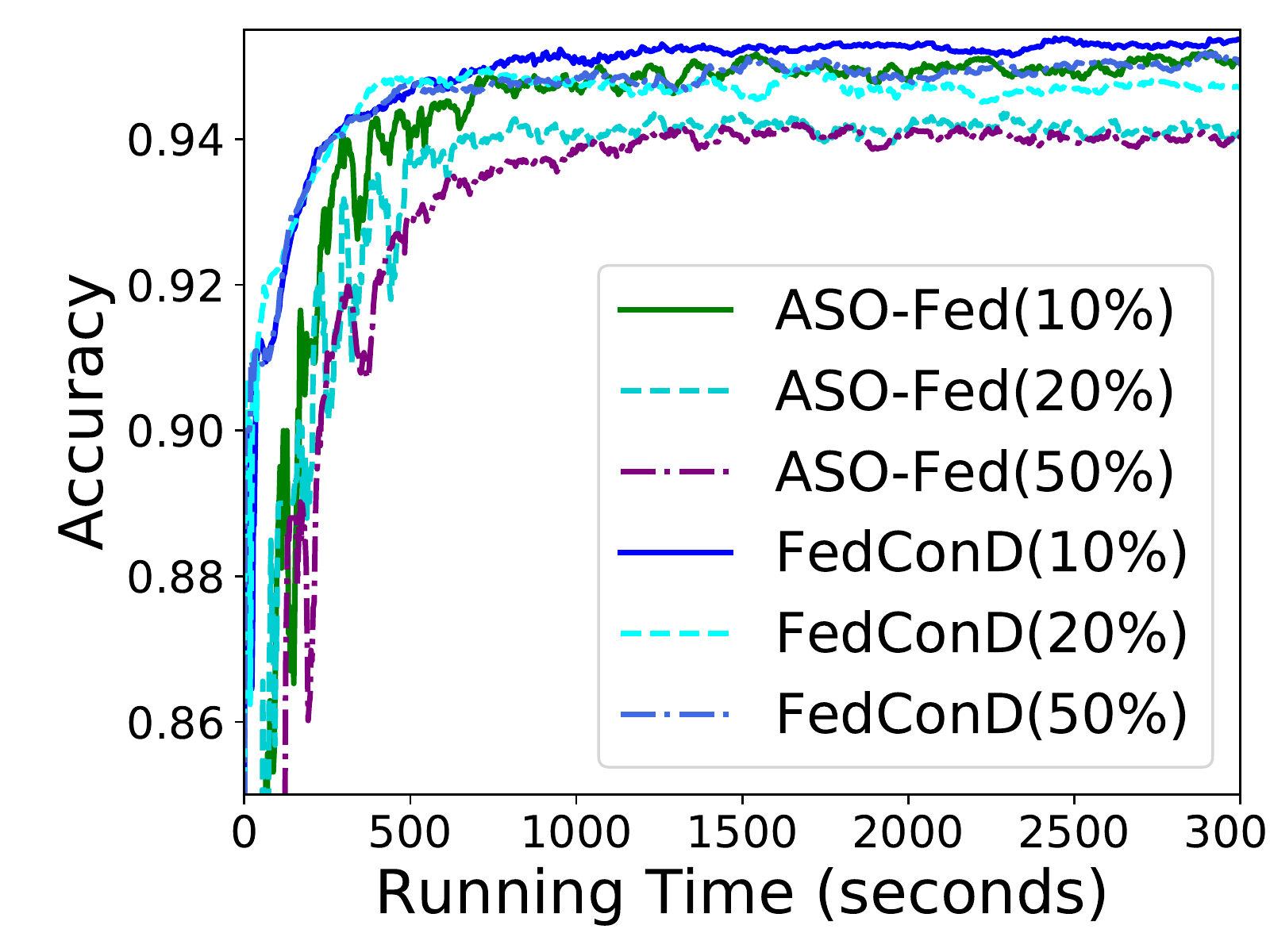}}
  \subfigure[Air Quality (SMAPE $\downarrow$)]{\includegraphics[scale=0.26]{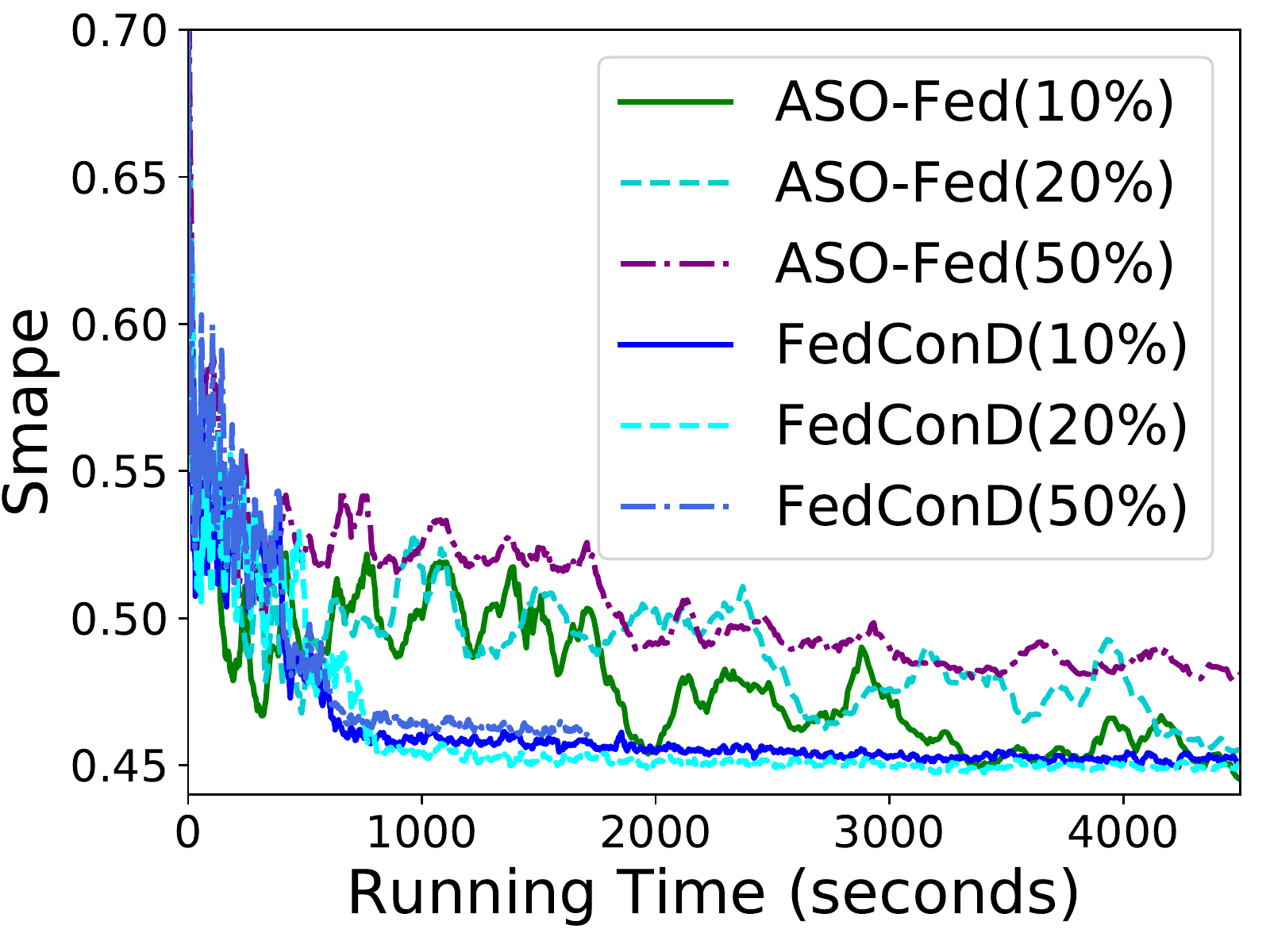}}
  \caption{Convergence of \model~compared with ASO-Fed on different portion of drift devices. }\label{tab:drift-percentage}
\end{figure}

\noindent \textbf{Concept Drift at Different Stages.} 

We perform another empirical study on model performance with drifts occurring  at different stages. Figure \ref{tab:drift-stage} shows the prediction performance of \model~compared with other FL algorithms for the sudden drift occurs at 10\% place (Early Stage), 40\% place (Middle Stage) and 70\% place (Late Stage) of the device data. 
ASO-Fed has larger errors than \model~when the drift happens at an early time, and this error gap shrinks as the drift occurs at later times. 
Synchronous FL algorithms have unstable prediction performance no matter when the drift starts. We also observe that the prediction errors drop as the drift starts late, which indicates a later drift affects the server model less.

\begin{figure}[htbp]
  \centering
  \subfigure[Fashion-MNIST (Accuracy $\uparrow$)]{\includegraphics[scale=0.26]{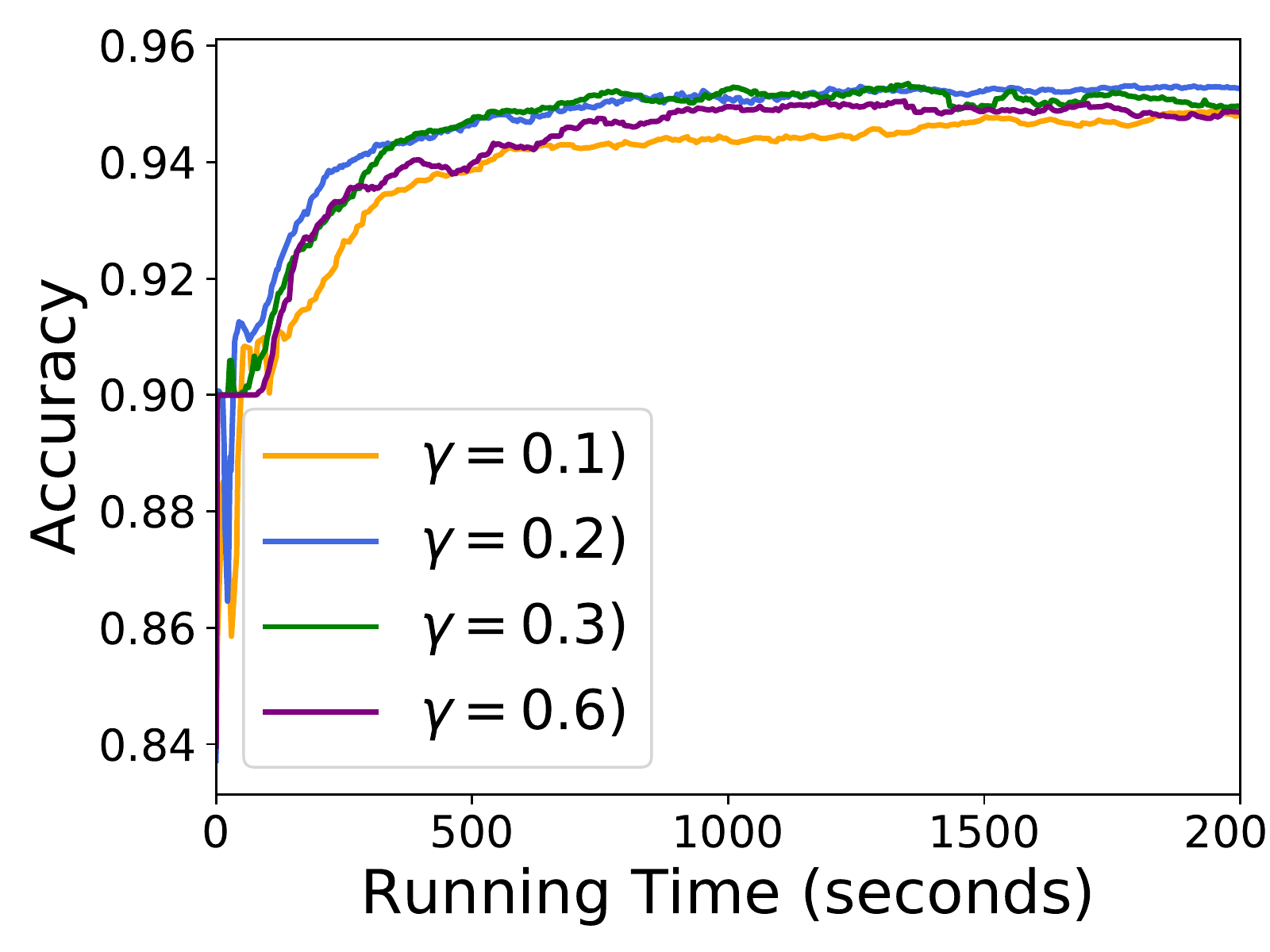}}
  \subfigure[FitRec (Smape $\downarrow$)]{\includegraphics[scale=0.26]{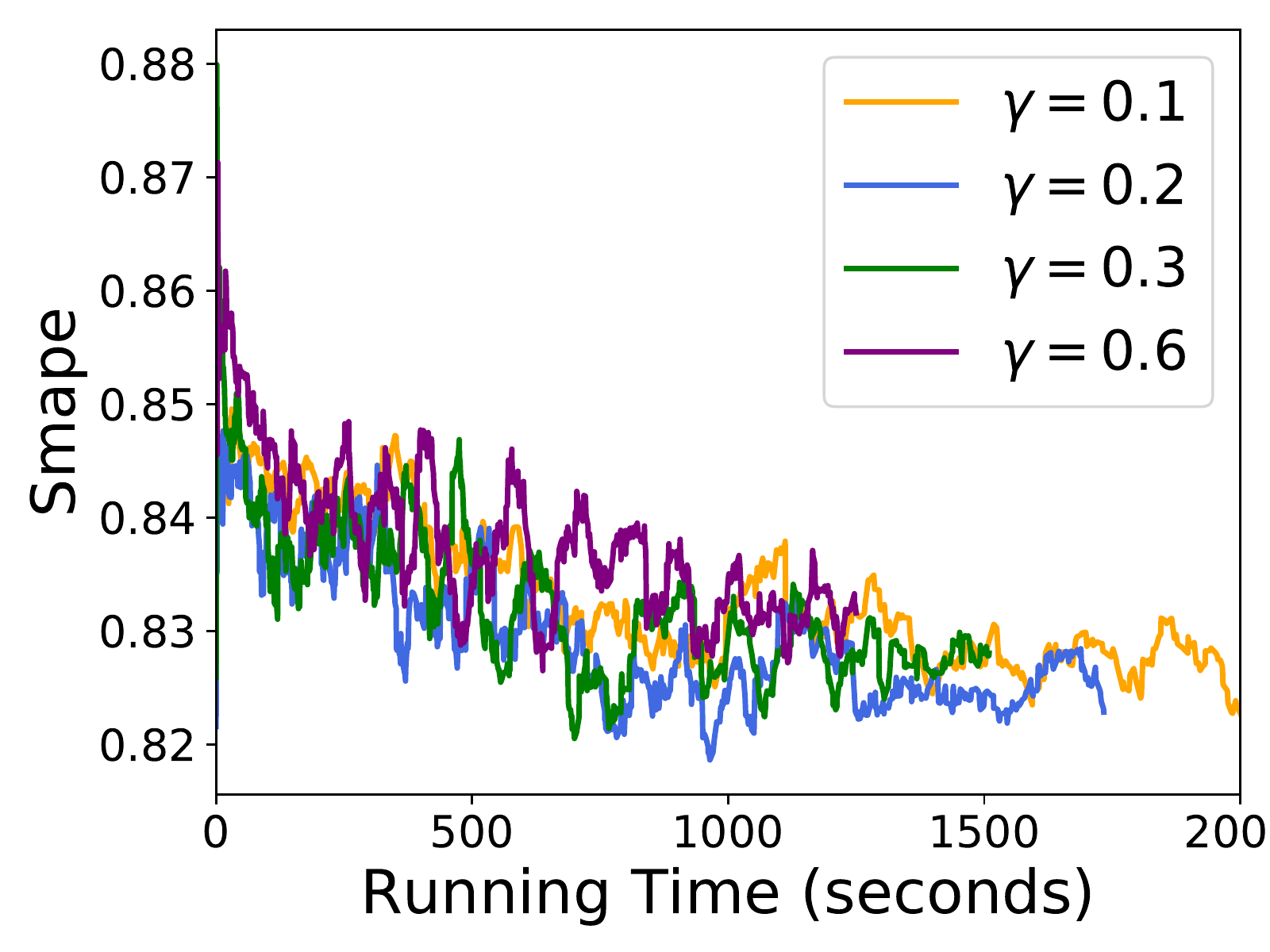}}
  \caption{For different fraction of local training devices ($\gamma$) simultaneously, the convergence of \model~varies. We increase $gamma$ from $0.1$ to $0.6$, the best model performance appears at $gamma = 0.2$.}\label{tab:model-gamma}
\end{figure}

\subsection{Sensitivity to Tunable Parameters}\label{tune-parameter}
\subsubsection{Number of Drift Devices} We set the portion of drift devices to be $10\%$ for the default experimental setting. In this section we also perform evaluations on the model performance with more devices which have concept drifts. As shown in Figure \ref{tab:drift-percentage}, we show the convergence of \model~compared to ASO-Fed with the number of drift devices increase. With $20\%$ drift devices, slight decrease in the prediction performance for  both approaches are observed for the Fashion-MNIST data. For the Air Quality data, the SMAPE value increases a little of ASO-Fed compared with $10\%$ drift devices setting, while there is almost no difference on the prediction performance of \model on $10\%$ and $20\%$ drift devices. We observe the similar situation for $50\%$ drift devices, where there are slight decrease on the prediction performance and the learning curves remain stable of \model. These results also prove the robustness of \model~on dealing with local concept drifts across an increasing number of devices.  

\subsubsection{Number of Local Training Devices Simultaneously} Asynchronous learning framework allows local devices to perform local training simultaneously, which is, local device can start its own training any time when it is ready. However, with all of local devices performing local training at the same time and upload the local models to the server, the server can easily become a bottleneck. Meanwhile, devices with better resources (e.g., network, hardware) will perform more updates to the server and lead to a biased global model. To deal with these problems, we design a parameter $\gamma$ to control the number of local active devices at the same time. 
Higher value of $\gamma$ implies more devices are performing the local training at the same time and lower value of $\gamma$ implies fewer devices perform local training simultaneously. 
In Figure \ref{tab:model-gamma}, we show the convergence trend of \model~with different $\gamma$ values. \model~achieves the best predictive performance when $\gamma = 0.2$ or $\gamma = 0.3$. However, with $\gamma = 0.2$, \model~converges faster, thus we use $\gamma = 0.2$ in the experiments of \model.

\begin{figure}[h]
  \centering
  \subfigure[Fashion-MNIST (Accuracy $\uparrow$)]{\includegraphics[scale=0.4]{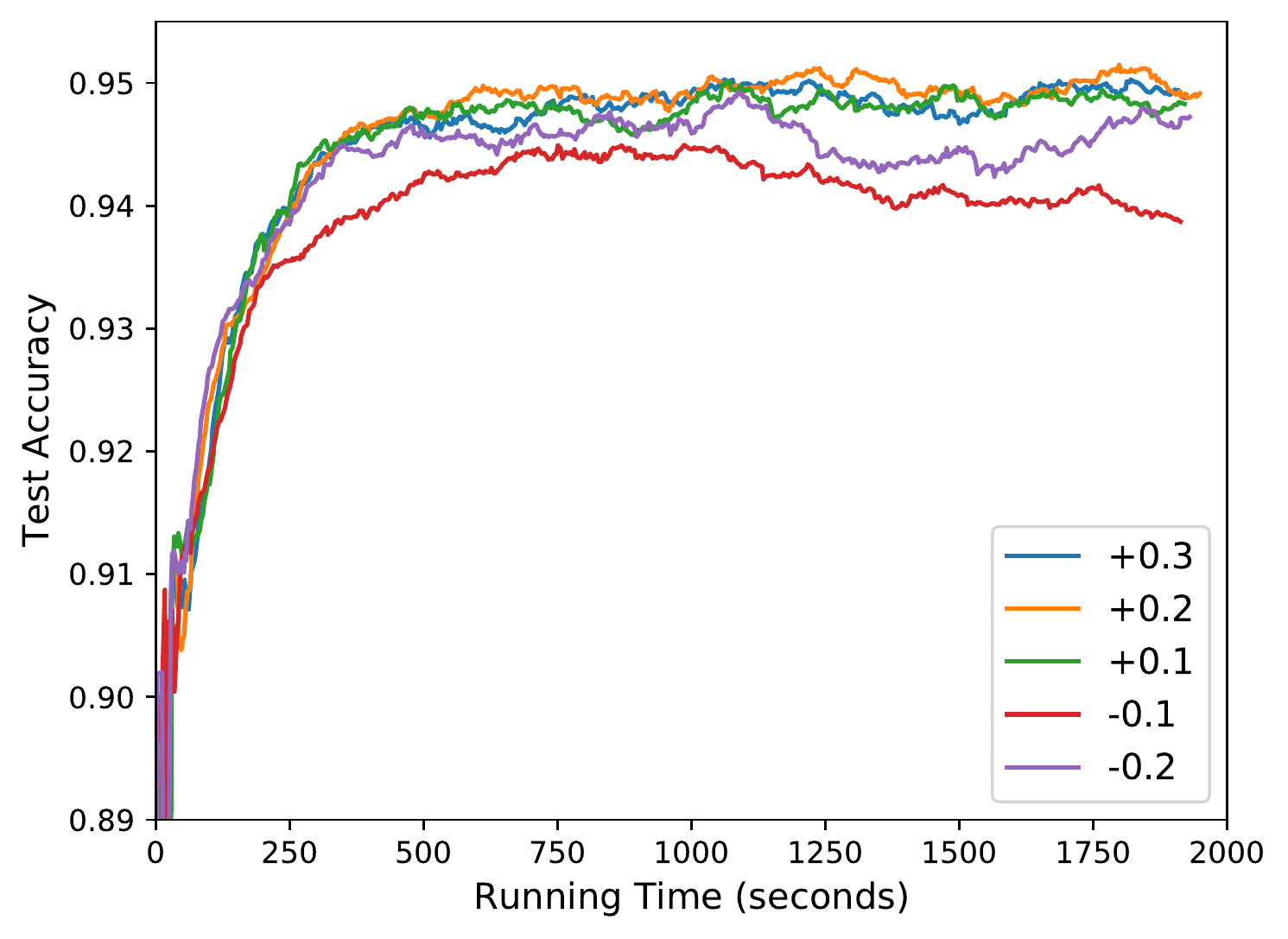}}
  \caption{Convergence trend of \model~on increasing or decreasing the regularization parameter $\lambda$ to handle concept drift. '+' means increasing $\lambda$ and '-' means decreasing $\lambda$.}\label{tab:model-lambda}
\end{figure}

\subsubsection{Regularization Parameter $\lambda$} 
Figure \ref{tab:model-lambda} displays the convergence trend of \model~for changing the penalty on local model with concept drifts. Intuitively, the local model will deviate further from the global model if concept drift occurs. These results confirms our claim in Section \ref{drift_handle}. When a concept drift is detected on local device, we increase the penalty to force the local model to be closer to the global model. In Figure \ref{tab:model-lambda}, the model performance is better when the regularization parameter, $\lambda$ is increased. Otherwise, the local model will deviate further from the global model and hurt the predictive performance on the global model. 

\section{Conclusion}
In this paper, we propose a novel framework, \model, to detect and handle concept drift in asynchronous federated learning. To the best of our knowledge, this is the first study of concept drift for the federated learning framework with heterogeneous device data.
On the server side, in order to get a more fair global model and reduce the overall communication cost, we design a strategy to balance the local updates and control the number of devices which perform local training at the same time. Experimental results on three evolving streaming data and two image data show that the proposed model can detect and handle concept drift in asynchronous federated learning efficiently.

\bibliographystyle{IEEEtran}
\bibliography{IEEEfull}


\end{document}